\documentclass{article}
\PassOptionsToPackage{nonamebreak}{natbib}
\usepackage{iclr2027_conference,times}
\usepackage{booktabs}
\usepackage{array}
\usepackage{multirow}
\usepackage{graphicx}

\usepackage{amsmath,amsfonts,bm}

\def\eqref#1{equation~\ref{#1}}

\def\1{\bm{1}}

\DeclareMathAlphabet{\mathsfit}{\encodingdefault}{\sfdefault}{m}{sl}
\SetMathAlphabet{\mathsfit}{bold}{\encodingdefault}{\sfdefault}{bx}{n}

\usepackage{hyperref}
\usepackage{url}
\usepackage{xcolor}
\usepackage{graphicx}
\usepackage{enumitem}
\usepackage{wrapfig}
\usepackage{subcaption}
\usepackage{placeins}

\newcommand{\modelname}{\textsc{LIFT}}
\makeatletter
\newcommand{\allowcitenamebreaks}{\def\NAT@nmfmt##1{{\NAT@up##1}}}
\makeatother
\AddToHook{env/table/begin}{\allowcitenamebreaks}
\AddToHook{env/table*/begin}{\allowcitenamebreaks}
\AddToHook{env/figure/begin}{\allowcitenamebreaks}

\title{Pretraining Latent Information Feedback Transformers with Teacher Supervision}

\author{
 \textbf{Dor Tirosh\textsuperscript{1}}\quad
 \textbf{Ido Amos\textsuperscript{2}}\quad
 \textbf{Mor Geva\textsuperscript{1}}
\\
 \textsuperscript{1}Blavatnik School of Computer Science and AI, Tel Aviv University
\\
 \textsuperscript{2}The Hebrew University of Jerusalem
\\
 \small\texttt{{\{dortirosh@mail,morgeva@tauex\}.tau.ac.il}}
}

\iclrfinalcopy
\begin{document}

\maketitle
\lhead{Preprint}

\begin{abstract}
Transformer language models (LMs) are feed-forward: deep-layer representations are never fed back to shallower layers, and the only pathway for information to flow downward across generation steps is the decoded token. This narrow channel forces models to recompute intermediate results and to discard alternative continuations. In this work, we remove this bottleneck \textit{during pretraining}, introducing the \modelname{} (Latent Information Feedback Transformer) architecture and training method which enable LMs to propagate state across generation.
We achieve this by turning recurrent-state learning into a teacher-forced prediction problem: each input token is paired with an information-dense state, derived from the next-token distribution of an off-the-shelf pretrained LM. The model, extended with a small number of additional parameters, is then trained to predict both the next token and the next state. As the input states are precomputed, pretraining remains fully parallel across positions.
At inference, the model's own predicted states are fed back, with a minor computational overhead that decreases with model size. 
Experiments with pretrained models ranging from 135M to 1B parameters show that \modelname{} consistently outperforms standard Transformers and baselines on language modeling, downstream reasoning tasks, and procedural tasks under token-matched budget, while being on par with or ahead of compute-matched Transformers.
Moreover, a controlled study on a state-tracking task shows that a tiny \modelname{} outperforms same-size Transformers trained on $8\times$ more data, even when trained with the states of a Transformer that fails the task.
Overall, we show that LMs can learn to exploit deep-to-shallow feedback during pretraining via scalable teacher supervision.\footnote{We release our code and trained models at \url{https://github.com/dortirosh1/LIFT/}.}
\end{abstract}

\section{Introduction}

Scaling language models (LMs) has been driven by major improvements to the Transformer architecture~\citep{fedus2022switch,deepseek2024v3,qiu2025gatedattention}. Still, the design is feed-forward: deeper layer representations are never fed back to shallower ones. The only pathway through which information propagates ``downward'' is the decoded token. This creates a narrow communication channel that bottlenecks the computation, leading the model to recompute previous computations~\citep{fan2021feedback,biran2024hopping} or drop possible continuations~\citep{zhu2025superposition}.

Several efforts have attempted to mitigate this bottleneck at post-training, by teaching models to generate compressed representations of their reasoning traces.
During inference, these methods propagate a continuous hidden representation at each step instead of a single token~\citep{hao2025coconut,shen2025codi,wei2025simcot}.
Although promising, such methods require expensive sequential latent rollouts during training, which may not suffice for rewiring the model to this new input distribution of richer continuous representations~\citep{rizvimartel2026illusion,wu2026llms}.

A natural question that arises is whether we can teach LMs to exploit such deep-to-shallow feedback \textit{at pretraining}, while avoiding infeasible training costs resulting from sequential processing.
Recent work tackles this by approximating full recurrence via Jacobi-style iterations, updating positions in parallel through multiple forward passes~\citep{zeng2026ponderlm2,cai2026t2mlr,wang2026fullbandwidth,huang2026latent,zhang2026whitematter}. However, this approach has an inherent quality--efficiency tradeoff: a better approximation requires more iterations and thus more training cost. 

\begin{figure}[t]
\setlength{\belowcaptionskip}{-8pt}
    \centering
    \includegraphics[width=0.95\linewidth]{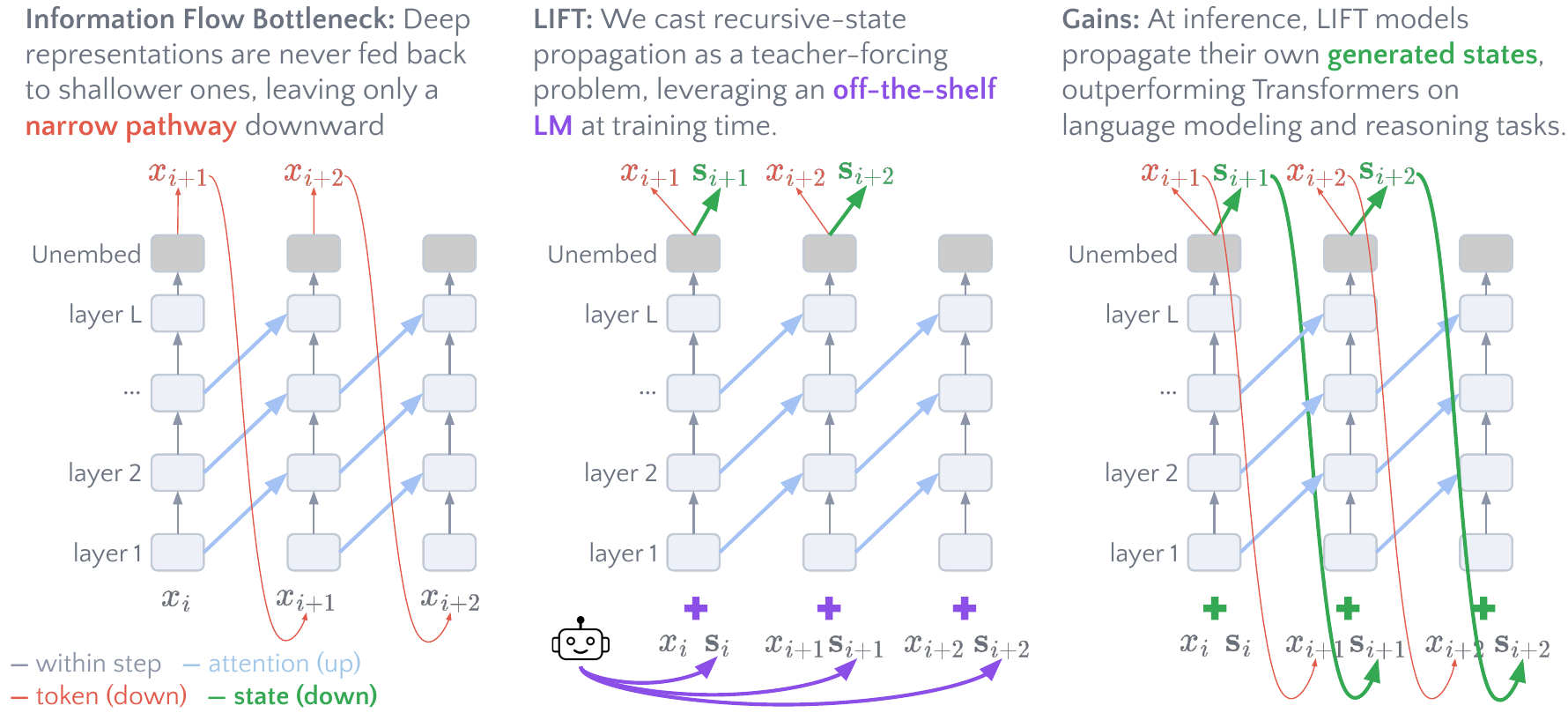}
    \caption{We remove the information flow bottleneck in Transformers (left) by introducing the LIFT architecture and pretraining approach (middle), which allows state propagation at inference (right).}
    \label{fig:intro}
\end{figure}

In this work, we propose \modelname{} (Latent Information Feedback Transformer), an alternative approach for pretraining LMs with state propagation, which converts recurrent-state learning into a teacher-forced prediction problem (Figure~\ref{fig:intro}).
During training, the model receives at every position a token \(x_i\) and state \(\mathbf{s}_i\), and is trained to predict both the next token \(x_{i+1}\) and the next state \(\mathbf{s}_{i+1}\).
To keep training efficient, we capitalize on the prevalence of publicly available LMs and obtain states
from \textit{next-token distributions} of an external ``teacher'' model\footnote{
Unlike in knowledge distillation~\citep{hinton2015distilling}, the teacher's role here is to supply the student's \emph{inputs}, not only its targets; at inference the teacher is removed and the student consumes its own states.} which runs in parallel on the same texts.
At inference, the model feeds its own predicted states back through this channel, alongside its generated tokens. Importantly, the states used in training are informative from the first training step and produced independently of the trained model, preserving parallelism across sequence positions.

We evaluate \modelname{} through comprehensive experiments, demonstrating its added expressivity and consistent performance gains.
First, we consider $S_5$ permutation composition, a state-tracking task that fixed-depth Transformers cannot solve as sequence length $N$ grows~\citep{merrill2024illusion}.
Training two-layer Transformer models on this task shows that, while a vanilla Transformer fails on sequences of $N\geq12$ steps, 
\modelname{} models solve 12-step sequences ($100\%$ accuracy) even when the states come from a Transformer teacher that itself fails at that length ($3\%$).
Thus, the feedback channel lets a fixed-depth model solve longer sequences than a Transformer of the same depth can, and parallel training on fixed-depth Transformer states is sufficient for learning it.

Next, we pretrain \modelname{} LMs from \(135\)M to \(1\)B parameters, on up to \(100\)B tokens at the \(1\)B scale~\citep[about \(5\times\) the Chinchilla budget;][]{hoffmann2022training}, and evaluate them on language modeling and a wide suite of downstream benchmarks.
Across evaluations, \modelname{} outperforms token-matched Transformers at every scale, as well as distillation and Jacobi-iteration feedback baselines, and is level with or ahead of compute-matched Transformers.
The largest gains are on procedural tasks, such as
multi-operand arithmetic and pattern continuation, where \modelname{} matches or beats a Transformer trained on $2{\times}$ more tokens at every scale.

Overall, our work shows that Transformer LMs can learn to exploit deep-to-shallow feedback through efficient, fully parallel pretraining with teacher-forced states, improving both language modeling and reasoning.
Our contributions can be summarized as follows:
\begin{itemize}
    \item We introduce, to our knowledge, the first method for pretraining a Transformer LM with a teacher-forced latent feedback channel: a pretrained LM supplies the states during training, keeping training fully parallel, and the model generates its own states at inference.
    \item We develop \modelname{}, which uses next-token distributions as states, aligning state supervision with the model's native prediction task.
    \item On a synthetic state-tracking task, we show that the feedback channel lets a fixed-depth \modelname{} solve the task at lengths where a Transformer of the same depth fails.
    \item We demonstrate the advantages of \modelname{} LMs of 135M to 1B parameters, which achieve better language modeling capabilities and downstream performance in token-matched and compute-matched settings, compared to a standard Transformer and to feedback variants. 
\end{itemize}

\section{Information Flow Bottleneck in Transformer LMs}
\label{sec:information-flow}

Consider an \(L\)-layer autoregressive Transformer LM, $p_\theta$, with hidden dimension $d$ and vocabulary \(\mathcal V\) \citep{vaswani2017attention}.
For an input sequence of tokens \(x_1, \dots, x_n\), let \(\mathbf h_i^\ell\in\mathbb R^d\) denote the representation at position \(i\)
after layer \(\ell\), with \(\mathbf h_i^0\) being the input token embedding. At each position, the output projection \(U\in\mathbb R^{|\mathcal V|\times d}\) produces the next-token distribution:
\begin{equation}
    \mathbf{p}_i = \operatorname{softmax}(\mathbf{o}_i) \;\; ; \;\; \mathbf{o}_i = U\mathbf h_i^L,
\label{eq:output_logits}
\end{equation}
where $\mathbf{o}_i$ are the logits.
At inference, information moves across positions via two channels: 
\begin{equation}
\begin{aligned}
\text{Attention channel:}\quad&
\mathbf h^\ell_{i \leq j} \bm{\rightarrow} \mathbf h_j^{\ell+1} \\
\text{Token feedback channel:}\quad&
\mathbf h_i^L \bm{\rightarrow} \mathbf p_i \rightarrow x_{i+1}
\bm{\rightarrow} \mathbf h_{i+1}^0.
\end{aligned}
\label{eq:mechanisms}
\end{equation}
Communication \textit{upward} is mediated by attention, which lets later positions \(j>i\) read the representation at a previous position \(i\).
Token feedback is the only way to move information \textit{downward}; a representation can influence 
a shallower hidden representation in subsequent steps \textit{only} at decoding time, and only by encoding information in the current position's output, which re-enters as input in the next step.
Crucially, this channel is extremely narrow, limited to a single decoded token \(x_{i+1}\), which can carry at most
\(\log_2|\mathcal V|\) bits of information
(\(\approx{17}\) bits for \(|\mathcal{V}|=100{,}000\)).

Allowing lower layers to access previously computed higher-layer representations, i.e., \(\mathbf{h}_i^L \bm{\rightarrow} \mathbf{h}_{i+1}^0\), could potentially enhance the model's computation.
First, later positions could reuse results computed in deep layers. Without such access, an intermediate result that emerges deep in the model is out of reach of lower layers that need it, so the model must recompute it earlier or fail~\citep{fan2021feedback}. For instance, when a model resolves the first hop of a multi-hop query ``too late'' for its remaining layers to complete the second, routing that late representation back into an earlier layer corrects up to \(66\%\) of the failures~\citep{biran2024hopping}.
Second, the model could carry several continuations forward, rather than committing to the sampled token. This matters when the model must explore alternatives: in graph reachability, continuous thoughts that hold many search paths at once let a two-layer Transformer find the answer in as many steps as the graph's diameter, instead of following a single path, which requires quadratically more steps~\citep{zhu2025superposition}.
Third, the computation's depth could grow with the sequence, as in a recurrent network. This matters for tasks whose number of sequential steps grows with the input, such as state tracking over a long sequence, which fixed-depth Transformers cannot solve~\citep{merrill2024illusion,merrill2025depth}.

\section{Latent Information Feedback Transformers (\modelname{})}
\label{sec:method}

Motivated by the problem above, we propose \modelname{}, a scalable architecture and training method for deep-to-shallow state propagation across positions, effectively removing the information flow bottleneck (Figure~\ref{fig:intro}).
We describe the modified architecture that fuses states with input tokens (\S\ref{sec:architecture}), and explain our approach for parallel training with teacher-forced states (\S\ref{sec:training}).

\subsection{\modelname{} Architecture for Next Token-State Prediction}
\label{sec:architecture}

Our goal is to let LMs propagate an additional state alongside the predicted token, carrying more information from deep layers. 
\modelname{} implements this by predicting a state along with every token:
\begin{equation}
    \label{eq:token-state}
    x_{i+1}, \mathbf{s}_{i+1} = p_\theta(\, x_{\leq i}, \mathbf{s}_{\leq i})
\end{equation}

\paragraph{Token and state prediction} In \modelname{} models, both $x_{i+1}$ and $\mathbf{s}_{i+1}$ are derived from the model's output logits $\mathbf{o}_i$. The next token $x_{i+1}$ is sampled as usual (Eq.~\ref{eq:output_logits}), while $\mathbf{s}_{i+1}$ is obtained by truncating the output distribution to its top-$k$ tokens: the $k$ largest logits are kept, the rest are set to $-\infty$, and a softmax with temperature $\tau$ renormalizes over the kept tokens:
\begin{equation}
    \label{eq:state}
    \begin{aligned}
        & \mathbf{o}'_i = \operatorname{top}_k(\mathbf{o}_i) \\
        & \mathbf{s}_{i+1} = \operatorname{softmax}(\mathbf{o}'_i / \tau) \in \mathbb{R}^{|\mathcal V|}, \\
    \end{aligned}
\end{equation}
so that $\mathbf{s}_{i+1}$ has at most $k$ non-zero entries. Building the state from the model's output distribution lets the model carry substantially more information than the sampled token: it ranks the alternative continuations and retains much of the context that produced it~\citep{morris2024language}.

Notably, a possible alternative for states could be the model's hidden representations before the output projection~\citep{hao2025coconut,shen2025codi,cai2026t2mlr}. We use distributions as they are the more natural choice in our case, where the model learns to predict the teacher's states (\S\ref{sec:training}). For any two models that share a tokenizer, these distributions live in the same space, where coordinates correspond to token probabilities, and predicting them is a well-established objective~\citep{hinton2015distilling}. The student's own predictions, which replace the teacher's states at inference, then live in that same space too. Hidden states have no shared coordinates across models~\citep{sriram2018cold}, so the student would have to regress onto another model's internal representation, which need not align with its own and which, in our early exploration, we found harder to predict.

\paragraph{Token and state fusion} To process the joint input of tokens and states, we design a dedicated \textit{fusion} layer that transforms them into a single representation, which feeds the standard Transformer blocks. Specifically,  $\mathbf{s}_i$ is first projected through the token embedding matrix $E \in \mathbb{R}^{|\mathcal V| \times d}$, yielding the empirical mean induced by $\mathbf{s}_i$, and then concatenated with $\mathbf{h}_i^0$. The concatenation is fused through a SwiGLU block~\citep{shazeer2020glu} with a residual connection to the token embedding:
\begin{equation}
    \label{eq:fusion}
    \begin{aligned}
        & \mathbf{u}_i = \big[\mathrm{RMSNorm}(\mathbf{h}_i^0) \;;\ \mathrm{RMSNorm}(W_s\, E^{\top} \mathbf{s}_i)\big] \\
        & \hat{\mathbf{h}}^0_i = \mathbf{h}_i^0 + W_d\big(\mathrm{SiLU}(W_g \mathbf{u}_i) \odot W_u \mathbf{u}_i\big) \\
    \end{aligned}
\end{equation}
where $W_s \in \mathbb{R}^{d \times d}$, $W_g, W_u \in \mathbb{R}^{2d \times 2d}$ and $W_d \in \mathbb{R}^{d \times 2d}$ are the fusion layer's parameters. At the first position, where no previous output exists, the projected state $E^{\top}\mathbf{s}_1$ is replaced by a learned vector. The fusion layer holds all the parameters \modelname{} adds: $11d^2$ weights and the initial-state vector, $3\%$ of our 1B model. Since the Transformer stack grows with $L d^2$, this share shrinks with depth. The fused representation $\hat{\mathbf{h}}^0_i$ is then fed to the standard Transformer stack, interacting with the context, which in turn contains information about previous tokens and states by the same mechanism.

\subsection{\modelname{} Training with Teacher-Forced States}
\label{sec:training}
Decoding in autoregressive Transformers is already sequential, so feeding back the state (Eq.~\ref{eq:token-state}) adds only a minor cost. However, at training time, such recurrent computations forfeit the parallelism that enables scalable training of Transformer LMs. To tackle this, we observe that this dependency disappears if the input states are computed \textit{in advance}: instead of obtaining them via autoregressive generation, the states can be supplied by an external teacher, through \textit{teacher forcing}.
This reduces the problem to finding a teacher capable of producing informative states. 
Here, we capitalize on the growing availability of open pretrained LMs~\citep{wolf2020transformers}, which offers many capable teachers, recycling their pre-invested compute
into a higher-quality training signal.

\paragraph{Training with teacher states}
Given a training sequence $\mathbf{x}=\langle x_1, \dots, x_n \rangle$, the teacher first processes $\mathbf{x}$ in parallel, emitting for every position $i$ its own output logits $\mathbf{o}^\star_i$.
These logits are then transformed according to Eq.~\ref{eq:state} to yield teacher states $\mathbf{s}^\star_{i+1}$. Notably, as the teacher is frozen, it admits common training optimizations that require fixed weights~\citep[e.g.,][]{xiao2023smoothquant,aminabadi2022deepspeedinference}.
Once the states are all available, they are paired with their matching tokens, i.e., $(x_{i+1}, \mathbf{s}^\star_{i+1})$. The pairs are then passed to the model's fusion layer \textit{in parallel} (Eq.~\ref{eq:fusion}), and forwarded through the full backbone as in standard causal-LM training, \textit{without} propagating gradients to the teacher. The state channel is thus teacher-forced, same as the token channel, and full parallelism is maintained.

At inference, states are produced by the model autoregressively, making input processing sequential. To allow prompt prefilling in one parallel pass without states, we employ \textit{prefix state dropout} during training: the states of each training sequence are replaced with probability \(p\) by a learned bias \(\mathbf{b}\in\mathbb{R}^d\) (\S\ref{app:prefix-dropout}). The model thus learns to operate both with and without states, allowing prefilling.

\paragraph{Mitigating train--inference state distribution shifts}
Because the model is trained on teacher states but consumes its own states at inference, a mismatch between the two distributions can accumulate over decoding steps~\citep{ranzato2016sequence}. We mitigate this in two ways. First, we supervise the model's state predictions with the teacher's, so that the states it generates at inference remain close to those it was trained on. To this end, we add a state-alignment term to the training objective:
\begin{equation}\label{eq:objective}
    \mathcal{L} = \sum_i \Big( \underbrace{-\log \mathbf{p}_i(x_{i+1})}_{\text{language modeling}} \;+\; \lambda\, \underbrace{\mathcal{D}_k\big(\mathbf{o}_i^\star \,\|\, \mathbf{o}_i\big)}_{\text{state alignment}} \Big),
\end{equation}
where \(\lambda \geq 0\) weights the two terms and \(\mathcal{D}_k\) is a forward-KL divergence~\citep{hinton2015distilling} adapted to top-\(k\) distributions (\S\ref{app:alignment-loss}).

Second, in the last \(10\%\) of training, we adapt the model to its own states, dropping the teacher and the alignment term.
Each training step executes one or two initial forward passes (with probability $50\%$), to produce the model's own states: the first pass reads the input tokens with the bias \(\mathbf{b}\) at every position. The second (optional) pass is fed the states predicted by the first. To train the model, we pass the states produced at the end of the process above instead of those typically supplied by the teacher, without propagating gradients back to the initial iterations (further details are in \S\ref{app:adaptation}).

\section{Evaluation of State Tracking Capabilities}
\label{sec:synthetic}

We start by evaluating \modelname{} on state tracking, showing the expressive advantage of its architecture over a standard Transformer, and how it can learn from both strong and weak teachers.

\begin{wrapfigure}{R}{0.33\linewidth}
\setlength{\belowcaptionskip}{-10pt}
    \centering
    \includegraphics[width=\linewidth]{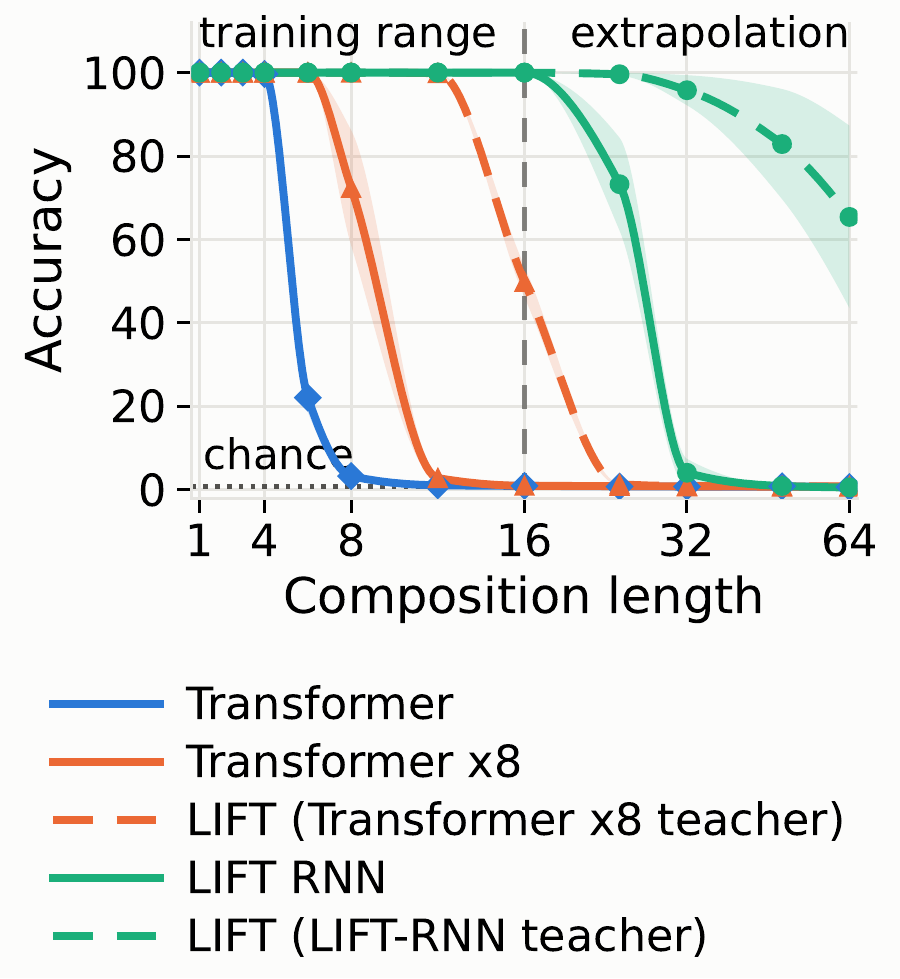}
    \caption{$S_5$ accuracy at increasing task lengths $N$ (linear for $N{\le}16$, logarithmic for $N{>}16$), of Transformers (solid lines) versus \modelname{} models (dashed lines). Standard error is over seeds.}
    \label{fig:s5}
\end{wrapfigure}
\paragraph{The $S_5$ task}
In this task, a model must predict the composition of a sequence of permutations, which requires state tracking along the sequence \citep{liu2023transformers}.
Let $S_5$ be the group of permutations of five elements (120 in total). Given an initial state $s_0 \in S_5$ and permutations $a_1,\dots,a_N \in S_5$, the model must predict the final state $s_N$ obtained by applying them to $s_0$:
\begin{equation}
    s_N = a_N \circ \cdots \circ a_1 \circ s_0, \qquad s_i = a_i \circ s_{i-1}.
    \label{eq:s5}
\end{equation}
We encode all the elements in $S_5$ as single tokens; thus, the model is given the sequence of tokens $\langle s_0, a_1, \dots, a_N \rangle$ and should predict a single token.
Importantly, this task is $\mathsf{NC}^1$-complete: unless $\mathsf{TC}^0 = \mathsf{NC}^1$, a Transformer of fixed depth cannot solve it for growing $N$, and the number of layers must grow as $\Omega(\log N)$~\citep{merrill2025depth}. Conversely, a recurrent model needs only a constant number of layers, as its computation depth grows with the sequence~\citep{merrill2024illusion}.

\paragraph{Experimental setup}
We train two-layer Transformers with 
$2$M parameters from scratch; \modelname{} adds its fusion layer ($7\%$ of the parameters). 
Input sequences of lengths $N \in [1,16]$ are drawn from $12$ fixed generators of $S_5$ at every training step, with a random initial state. We report accuracy on $2{,}000$ held-out sequences per length, up to $N{=}64$, averaged accuracy over 3 seeds (see \S\ref{app:s5-setup} for more details).
We compare the following models:
\begin{enumerate}
[leftmargin=*,itemsep=2pt,topsep=3pt,parsep=0pt]
    \item \textbf{Transformer}: A token-matched vanilla Transformer, trained for $5{,}000$ steps. Following \citet{liu2023transformers}, the loss is computed on the correct state $s_i$ at every position $i$, not only on the final $s_N$. At test time, it answers in one parallel pass.

    \item \textbf{Transformer$\times$8}: Same as the above baseline, except that the model is trained $8\times$ longer ($40{,}000$ steps). Notably, larger training budgets do not yield substantial improvements (see \S\ref{app:transformer-budget}).
    
    \item \textbf{\modelname{}-RNN}: The \modelname{} architecture trained sequentially for $40{,}000$ steps: at every position the output distribution is embedded with the token and passed to the next position, and gradients flow across steps using only the answer loss.

    \item \textbf{\modelname{} (Transformer$\times$8 teacher)}: Trained in parallel for $5{,}000$ steps with this teacher.
    
    \item \textbf{\modelname{} (RNN teacher)}: Trained in parallel for $5{,}000$ steps with \modelname{}-RNN as a teacher.
\end{enumerate}

\paragraph{Results}
Figure~\ref{fig:s5} shows that the fixed-depth Transformer struggles with this task, as theory predicts: even with $8\times$ the training budget, it solves only lengths up to $N{=}6$ and is at chance from $N{=}12$ on. Training sequentially with latent feedback (\modelname{}-RNN) removes this bottleneck, allowing a model of the same depth to solve every length in the training range. When trained fully in parallel on the states of this recurrent model, \modelname{} inherits its unbounded computation depth: it matches the teacher in the training range and even extrapolates beyond it, achieving $96\%$ at $N{=}32$ and $65\%$ at $N{=}64$, although trained only on lengths up to $N{=}16$. 
Interestingly, \modelname{} shows such extrapolation even when trained with a depth-bounded non-recurrent teacher.
\modelname{} solves every length up to $N{=}12$ and half of the sequences at $N{=}16$, whereas its teacher, a same-size Transformer trained with $8\times$ the budget, is at chance beyond $N{=}8$.
We stress that, trained this way, \modelname{} does not acquire unbounded depth and fails beyond the training range. Still, learning from the teacher's states yields a large gain over a standard Transformer, even one trained much longer. 

\section{Experiments with Pretrained LMs}
\label{sec:experiments}

We conduct comprehensive evaluations of \modelname{} LMs in two setups, benchmarking against vanilla Transformer and distillation baselines on a wide set of benchmarks (\S\ref{sec: results}), and comparing with feedback Transformers that approximate recurrent states through repeated parallel passes (\S\ref{sec:jacobi}).

\subsection{Experimental Setting}
\label{sec:setting}

\paragraph{Pretrained LMs}
We pretrain \modelname{} LMs with 135M, 350M and 1B parameters, using the OLMo~2 architecture, codebase and pretraining data~\citep{olmo2024olmo2}, each for $5\times$ its Chinchilla token budget~\citep{hoffmann2022training}. Models at 135M scale are trained with three seeds, and larger models with one seed (due to compute constraints).
The teacher at each scale is a same-size model pretrained for $10\times$ its Chinchilla budget; stronger teachers, either larger or trained for much longer, change the results only marginally (\S\ref{app:teacher-strength}).
\modelname{} uses $k{=}1024$, $\tau{=}1.5$ and $\lambda{=}1.5$ at every scale; see \S\ref{app:ablations} for ablations and \S\ref{app:impl} for additional implementation details. We compare against the following baselines, all evaluated after an identical learning-rate annealing phase~\citep{wen2024wsd}:
\begin{enumerate}
[leftmargin=*,itemsep=2pt,topsep=3pt,parsep=0pt]
    \item \textbf{Vanilla Transformer}: A standard OLMo~2 Transformer, identical to \modelname{} except for the fusion layer. 
    We report results for a model trained on the same token budget and on the same compute budget as ours, accounting for the teacher's forward pass (\S\ref{app:compute-matched}).
    \item \textbf{Distillation}: The vanilla Transformer trained with the forward-KL distillation term~\citep{hinton2015distilling} in Eq.~\ref{eq:objective}, isolating the effect of the distillation objective from that of the latent feedback.
    \item \textbf{\modelname{} w/o states}: Our trained \modelname{} LM with the channel ``switched off'': in place of a state, the fusion layer receives the learned bias $\mathbf{b}$ of prefix state dropout (\S\ref{sec:training}). No state is fed back, so the model runs as a standard Transformer in one parallel pass.
\end{enumerate}

\paragraph{Comparison with feedback Transformers}
We adapt the setup of T\textsuperscript{2}MLR \citep{cai2026t2mlr}: a Transformer whose layers read a state computed by deeper layers at earlier positions, pretrained by refining all positions in parallel over repeated passes (Jacobi iterations) instead of sequentially. All models share the SmolLM2-135M backbone~\citep{allal2025smollm2smolgoesbig} and are trained 
on a 10B-token FineWeb-Edu sample~\citep{penedo2024finewebdatasetsdecantingweb}. We compare \modelname{} with \textbf{T\textsuperscript{2}MLR}, both our own training run and the published checkpoint. In addition, we compare with the concurrent models of \citet{wang2026fullbandwidth}, which feed the hidden state at the previous position back to the input and train with a few passes per token. At the time of writing, the official code and models are not publicly available, so we implemented the method ourselves, using the schedule of the authors' larger runs; we call this adaptation \textbf{Multi-pass Transformer}.
We also compare against a vanilla Transformer. 
Models are compared under (a) matched tokens, each trains on about 10B tokens, and (b) matched compute, each receives the training FLOPs of T\textsuperscript{2}MLR's 10B-token run, so cheaper methods train on more tokens, revisiting the data once exhausted. See \S\ref{app:jacobi-setup} for implementation details.

\paragraph{Evaluation benchmarks}
We use the following evaluations for both sets of experiments:
\begin{enumerate}
[leftmargin=*,itemsep=2pt,topsep=3pt,parsep=0pt]
    \item \textbf{Language modeling}: Perplexity on held-out text of the training corpus: for the three-scale experiments, the 1M-token C4~\citep{raffel2020exploring} validation subset of the OLMo~2 evaluation set~\citep{olmo2024olmo2}; for the feedback comparison, a held-out FineWeb-Edu sample of 127K tokens. Perplexity on the other sources of the OLMo~2 evaluation set is reported in \S\ref{app:additional_results}.
    \item \textbf{Downstream tasks}: Twenty benchmark suite run through OLMES~\citep{gu2024olmes}, covering the base-model evaluation families of OLMo~2 and OLMo~3~\citep{olmo2025olmo3}. Eight of these are at chance or floor (under 3\%) for all models at these scales; 
    the other twelve form three groups, each reported as the mean over its benchmarks (the full list is in \S\ref{app:benchmarks}):
    \begin{itemize}[leftmargin=*,itemsep=1pt,topsep=1pt,parsep=0pt]
        \item \emph{MC}: Multiple-choice questions on science, commonsense and world knowledge. 
        Each answer option is scored by its likelihood given the question and the highest wins.
        \item \emph{Gen.}: Generative tasks on reading comprehension, knowledge recall and multi-step reasoning.
        The model generates the answer, scored by F1 or exact match against the reference.
        \item \emph{RC}: Reference-completion tasks in science, knowledge, math and code.
        The reference answer is scored by its likelihood under the model, reported in bits-per-byte (BPB), lower is better.
    \end{itemize}
    \item \textbf{Arithmetic}: Our own synthetic task, testing multi-step computation carried in the model's latent state (no chain of thought). Each sample adds and subtracts 2--10 single-digit numbers, e.g.\ \texttt{2 + 0 - 6 + 7 + 5 =} (answer \texttt{8}); we report the correct answer's BPB (more metrics in \S\ref{app:benchmarks}).
\end{enumerate}

\subsection{Evaluating Language Modeling, Knowledge, and Reasoning Capabilities}
\label{sec: results}

\paragraph{\modelname{} outperforms Transformer baselines at matched tokens and at matched compute}
Table~\ref{tab:lm} presents the main results, with a per-benchmark breakdown in \S\ref{app:additional_results}. At matched tokens, \modelname{} is ahead on every metric at every scale: perplexity drops by \(5\)--\(5.5\%\), and it wins 33 of the 36 downstream benchmark comparisons. At matched compute, where the Transformer trains on over \(40\%\) more data, \modelname{} keeps the lead on every metric at 350M and 1B and is level with or ahead of it at 135M. These gains are consistent over the three 135M-scale training runs (see \S\ref{app:res-seeds}).
Importantly, the gains come from the added channel rather than from the training signal alone:
\modelname{} w/o states, the same weights run without the channel, and the distillation baseline, trained with the same state-alignment objective but without the channel, trail \modelname{} on perplexity and on most of the benchmarks at every scale. 
Further ablations reinforce this (\S\ref{app:ablations}): training \modelname{} with a narrower channel, without the state-alignment objective, or without the teacher at all significantly erodes the gains.

\begin{figure}[t]
    \centering
    \begin{subfigure}[t]{0.49\linewidth}
        \centering
        \includegraphics[width=\linewidth]{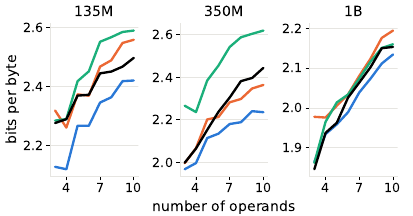}
        \caption{Arithmetic}
        \label{fig:arith_bpb}
    \end{subfigure}
    \hfill
    \begin{subfigure}[t]{0.49\linewidth}
        \centering
        \includegraphics[width=\linewidth]{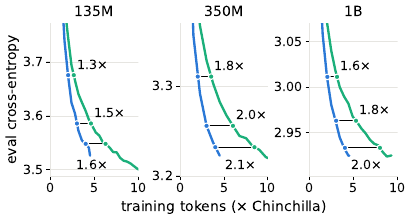}
        \caption{Language Modeling}
        \label{fig:ce_multiplier}
    \end{subfigure}
    \\[1mm]
    \includegraphics[width=\linewidth]{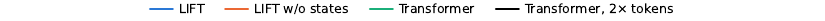}
    \caption{\modelname{} performance on arithmetic and language modeling. \textbf{(a)} \modelname{} shows a consistent advantage on arithmetic that persists as task complexity increases, outperforming a baseline trained on $2\times$ more tokens. \textbf{(b)} Throughout training, \modelname{} is more token-efficient than a vanilla Transformer, and the advantage \textit{grows} with training: each horizontal segment marks the additional tokens the Transformer needs to reach the same loss.}
    \label{fig:lm_results}
    \vspace{-1mm}
\end{figure}

\paragraph{\modelname{}'s token advantage grows with training}
Figure~\ref{fig:ce_multiplier} compares the cross-entropy on the evaluation set at matched checkpoints along training,
showing how many more training tokens the Transformer baseline needs to reach the performance of \modelname{}. At every scale, this multiplier grows along training: the longer both models train, the more tokens the Transformer needs to catch up, suggesting that the token efficiency of \modelname{} increases with training.
\paragraph{The largest gains are on procedural tasks}
Across all evaluations, \modelname{} achieves the largest gains on procedural tasks---our arithmetic task and the pattern continuation task from the OLMES basic-skills suite---which require a step-by-step answer derivation. 
On arithmetic, \modelname{} is ahead of the Transformer baseline at every scale, outperforming its own teacher, which was trained on $2\times$ more tokens (\(1\)--\(5\%\) fewer bits on average; \S\ref{app:res-arith}). These gains persist across increasing numbers of operands (Figure~\ref{fig:arith_bpb}). Pattern continuation asks for the next element of a short sequence that follows a rule, over numbers, letters, symbols or dates (e.g., \texttt{2 4 6 8} \(\to\) \texttt{10}, \texttt{E G I K} \(\to\) \texttt{M}). \modelname{} beats the token-matched Transformer by \(6\)--\(10\%\) in bits per byte at every scale, and the \(2\times\)-token Transformer by \(4\)--\(5\%\) at 350M and 1B (a tie at 135M; \S\ref{app:res-pattern}).

\paragraph{Performance with parallel prefill}
\modelname{} models can process the input prompt sequentially (with state predictions) or in parallel (with the bias state; \modelname{} w/o states in Table~\ref{tab:lm}).
The results show that \modelname{} w/o states is comparable to the token-matched Transformer baseline across all scales, indicating \modelname{} maintains strong capabilities without sequential processing of the input. In \S\ref{app:prefill}, we compare additional methods for performing parallel prefill with \modelname{} models, which vary in efficiency and quality. We find that iteratively refining the prompt, as in the adaptation phase (\S\ref{sec:training}), recovers much of the gain of sequential processing within a few iterations.

\begin{table}[t]
\centering
\small
\setlength{\tabcolsep}{4pt}
\renewcommand{\arraystretch}{1.0}
\caption{Performance of \modelname{} LMs against non-feedback Transformer baselines at three scales.}
\label{tab:lm}
\begin{tabular*}{\linewidth}{@{\extracolsep{\fill}}clrccccc@{}}
\toprule
 & Model & Tokens & PPL $\downarrow$ & Arith.\ {\scriptsize BPB} $\downarrow$ & MC acc.\ \% & Gen.\ {\scriptsize F1/EM} \% & RC {\scriptsize BPB} $\downarrow$ \\
\midrule
\multirow{6}{*}{\rotatebox[origin=c]{90}{135M}} & \modelname{} & 13.4B & \textbf{30.65} & \textbf{2.29} & 39.4 & \textbf{15.5} & \textbf{1.078} \\
 & \modelname{} w/o states & 13.4B & 32.44 & 2.42 & 38.7 & 12.0 & 1.105 \\
 & Transformer, token-match & 13.4B & 32.43 & 2.46 & 38.1 & 13.6 & 1.110 \\
 & Transformer, compute-match & 19.3B & 31.04 & 2.50 & \textbf{39.5} & 14.1 & 1.080 \\
 & Distillation & 13.4B & 31.77 & 2.36 & 39.1 & 12.6 & 1.089 \\
\midrule
\multirow{6}{*}{\rotatebox[origin=c]{90}{350M}} & \modelname{} & 34.9B & \textbf{23.82} & \textbf{2.12} & \textbf{45.0} & \textbf{23.2} & \textbf{0.921} \\
 & \modelname{} w/o states & 34.9B & 24.96 & 2.21 & 44.2 & 21.0 & 0.945 \\
 & Transformer, token-match & 34.9B & 25.15 & 2.46 & 43.2 & 21.0 & 0.956 \\
 & Transformer, compute-match & 49.9B & 24.32 & 2.33 & 44.1 & 22.0 & 0.942 \\
 & Distillation & 34.9B & 24.52 & 2.23 & 44.5 & 22.1 & 0.932 \\
\midrule
\multirow{6}{*}{\rotatebox[origin=c]{90}{1B}} & \modelname{} & 107.4B & \textbf{18.81} & 1.99 & \textbf{52.5} & \textbf{35.0} & \textbf{0.772} \\
 & \modelname{} w/o states & 107.4B & 19.62 & 2.05 & 51.2 & 32.9 & 0.791 \\
 & Transformer, token-match & 107.4B & 19.79 & 2.03 & 50.5 & 32.9 & 0.808 \\
 & Transformer, compute-match & 152.4B & 19.17 & 2.07 & 51.6 & 34.8 & 0.792 \\
 & Distillation & 107.4B & 19.40 & \textbf{1.95} & 52.0 & 34.8 & 0.781 \\
\bottomrule
\end{tabular*}
\end{table}

\subsection{Comparison with Jacobi-Iteration Feedback Transformers}
\label{sec:jacobi}
Table~\ref{tab:jacobi} shows the average results over three training seeds, with seed noise and additional results reported in \S\ref{app:jacobi-results}. At matched tokens, \modelname{} shows the best average performance on every metric, tied with T\textsuperscript{2}MLR on generation tasks, though we note that in this setting all methods are within noise.
At matched compute, \modelname{} is on par with or better than the vanilla Transformer on all tasks, outperforming the latent feedback baselines. On the arithmetic task, performance is within the noise estimate compared to our reproduction of T\textsuperscript{2}MLR, though this is mainly due to the relatively high variance of the latter. Notably, the feedback baselines fall behind the vanilla Transformer on every metric (except arithmetic) once compute is matched.

Crucially, pretraining with multiple iterations introduces a tradeoff between approximation quality and training efficiency. More iterations improve state approximation, providing a better training signal per token, but they make each token more expensive to process \citep{cai2026t2mlr}. 
Our results reflect this tradeoff: with 16 refinement passes, T\textsuperscript{2}MLR achieves lower perplexity than the Multi-pass Transformer at matched tokens, but at matched compute, their order reverses as the Multi-pass Transformer's fewer passes let it train on more tokens. \modelname{} obtains its training states from a teacher, avoiding this iterative approximation, and is equal to or better than both methods on every metric under either budget.
Multi-pass training also increases memory use because activations from the passes used for backpropagation must be stored. \modelname{} uses less peak memory than both feedback baselines at the measured batch sizes (see \S\ref{app:jacobi-footprint} for memory usage comparison).

\begin{table}[t]
\centering
\small
\setlength{\tabcolsep}{3.4pt}
\renewcommand{\arraystretch}{1.0}
\caption{Comparison with Jacobi-iteration feedback Transformers at matched tokens and matched compute. We report results for T\textsuperscript{2}MLR using the official checkpoint (\textit{``HF''}) and our reproduction.}
\label{tab:jacobi}
\resizebox{\linewidth}{!}{%
\begin{tabular}{@{}lrrrrrr@{\hspace{14pt}}rrrrrr@{}}
\toprule
& \multicolumn{6}{c}{\textbf{Token-matched} \textcolor{gray}{10B tokens}} & \multicolumn{6}{c}{\textbf{Compute-matched} \textcolor{gray}{$2.3{\times}10^{19}$ FLOPs}} \\
\cmidrule(lr){2-7}\cmidrule(l){8-13}
Model & FLOPs & PPL $\downarrow$ & Arith.\ $\downarrow$ & MC \% & Gen.\ \% & RC $\downarrow$ & Tokens & PPL $\downarrow$ & Arith.\ $\downarrow$ & MC \% & Gen.\ \% & RC $\downarrow$ \\
\midrule
\modelname{} & 1.44e19 & \textbf{19.42} & \textbf{2.18} & \textbf{36.7} & \textbf{7.8} & \textbf{1.705} & 16.2B & 18.04 & \textbf{2.18} & 38.0 & \textbf{9.5} & 1.629 \\
Transformer & 1.02e19 & 20.30 & 2.35 & 36.2 & 6.7 & 1.761 & 23.0B & \textbf{18.02} & 2.30 & \textbf{38.1} & 8.5 & \textbf{1.622} \\
T\textsuperscript{2}MLR (our run) & 2.32e19 & 19.92 & 2.27 & 36.5 & \textbf{7.8} & 1.737 & 10.0B & 19.92 & 2.27 & 36.5 & 7.8 & 1.737 \\
T\textsuperscript{2}MLR (HF) & 2.35e19 & 21.55 & 2.38 & 35.3 & 6.5 & 1.816 & 10.1B & 21.55 & 2.38 & 35.3 & 6.5 & 1.816 \\
Multi-pass Transformer & 1.31e19 & 20.40 & 2.25 & 36.3 & 6.5 & 1.773 & 17.8B & 18.58 & 2.25 & 37.6 & 7.7 & 1.675 \\
\bottomrule
\end{tabular}}
\end{table}

\section{Related Work}
\label{sec:related}

\paragraph{Latent feedback in post-training}
A growing body of work attempts to post-train LMs to use forms of latent feedback, typically focusing on reasoning tasks. \citet{hao2025coconut} train LMs to reason by treating the last hidden state as a latent token, processed for a fixed number of steps. This approach, and several follow-up efforts that adapt the learning signal~\citep{shen2025codi,wei2025simcot}, rely on sequential rollouts that are difficult to scale. To reduce this cost, \citet{wu2025pccot} replace the sequential rollouts with Jacobi iterations. Rather than replacing tokens with latents, \citet{tang2026turbo} keep the decoded token and route the previous token's higher-layer states to lower layers, fine-tuning the model with sequential processing over groups of tokens. Alternative approaches feed back soft tokens, the probability-weighted average of token embeddings, similar to \modelname{}. These soft tokens are used either without any training~\citep{zhang2025soft,zhuang2025mixture} or with reinforcement learning~\citep{butt2026soft}. Finally, some works teacher-force the latent inputs using supervision from gold reasoning traces~\citep{tan2025colar,amos2026thinkingstates} or from an oracle~\citep{gozeten2026continuous}; this enables parallel training but requires annotated reasoning.

\paragraph{Latent feedback in pretraining}
More related to our work are methods focusing on latent feedback during pretraining. Early approaches compute the states with exact recurrence, which makes training sequential over positions~\citep{fan2021feedback} or over segments~\citep{bulatov2022recurrent}. More recent feedback LMs instead approximate Jacobi iterations, following fixed-point iteration methods developed to parallelize RNN training~\citep{lim2024parallelizing,danieli2026pararnn}. These LMs feed back either keys and values~\citep{zhang2026whitematter} or hidden states~\citep{zeng2026ponderlm2,huang2026latent,cai2026t2mlr,wang2026fullbandwidth}, and trade training cost for approximation quality, a tradeoff we examine in \S\ref{sec:jacobi}. In all of these methods, states are self-generated at training time, requiring sequential rollouts or an approximation, unlike \modelname{}, which teacher-forces the states with a trained LM. In \citet{kumar2026pretraining}, an RNN is similarly pretrained with teacher-forced memory, but using an encoder trained in tandem. \citet{tack2026llm} use a pretrained teacher, but only as a source of targets: the model predicts concepts extracted from it and mixes them into its hidden states. We provide an extended discussion and detailed comparison in \S\ref{app:related}.

\section{Conclusion}
We introduce \modelname{}, a Transformer LM enhanced with deep-to-shallow feedback during pretraining, realized by a state constructed from the next-token output distribution. Unlike prior methods that rely on expensive sequential rollouts or state approximations, \modelname{} leverages an external teacher model to provide informative states, supporting parallel training. During pretraining, \modelname{} is aligned to predict its own states, removing the reliance on the teacher at inference time. In a controlled setting on an $S_5$ state-tracking task, a two-layer \modelname{} successfully solves the task at lengths where its own teacher, trained for $8\times$ longer, fails to exceed chance performance. Through extensive LM pretraining experiments at the 135M--1B parameter scale, we show \modelname{} outperforms a standard Transformer on a variety of tasks, when matched in tokens and compute, and displays training token efficiency that grows as training progresses. Compared against alternative feedback methods, \modelname{} is comparable to or better than all baselines considered in both token and compute matched setting.

Notably, our work focuses on a certain choice for the teacher that shares the architecture of the student. Studying the effects of teachers with other sizes and possibly different vocabularies is an attractive avenue for future work. Another is training \modelname{} far beyond its teacher's training budget, a regime in which its performance and training dynamics remain open questions.

\subsubsection*{Acknowledgments}
We thank Oren Pereg and Jonathan Mamou for their support and useful discussions in early stages of this work.
We also thank Ziyang Cai and Xingyu Zhu for kindly sharing their trained T\textsuperscript{2}MLR models.
This research was supported in part by Intel and a Moonshot Grant by the Planning and Budgeting Committee of the Council for Higher Education in Israel. 
It was made possible through a GPU compute resource grant funded by the Association of University Heads, the Council for Higher Education and the AI Research Compute Center.

\bibliography{iclr2027_conference}
\bibliographystyle{iclr2027_conference}

\appendix
\raggedbottom
\section{Additional Method Details}
\label{app:method-details}

\subsection{Prefix State Dropout}
\label{app:prefix-dropout}
For each training sequence, with probability \(p\), we draw a prefix length \(m\sim\mathcal{U}\{0,\dots,n\}\) and feed the fusion layer (Eq.~\ref{eq:fusion})
\begin{equation}\label{eq:prefix-dropout}
    \tilde{\mathbf{e}}_i =
    \begin{cases}
        \mathbf{b} & 1 < i \leq m \\
        E^{\top}\mathbf{s}^\star_i & i > m
    \end{cases}
\end{equation}
in place of the projected state \(E^{\top}\mathbf{s}^\star_i\). The first position always keeps the learned initial-state vector, since no previous output exists there at inference either. Because \(m\) ranges up to the full length, the model also learns to process an entire sequence without states, the mode evaluated as \modelname{} w/o states (\S\ref{sec: results}). The same augmentation is applied to the loss-carrying pass of the adaptation phase (\S\ref{app:adaptation}).

\subsection{The State-Alignment Loss}
\label{app:alignment-loss}
The alignment term \(\mathcal{D}_k\) of Eq.~\ref{eq:objective} compares the teacher's and the student's distributions at the state temperature \(\tau\) on the \(k\) tokens of \(\mathbf{s}^\star_{i+1}\), aggregating the remaining probability mass into a single tail outcome.
Formally, let \(S\) be the \(k\) tokens of \(\mathbf{s}^\star_{i+1}\), let \(\mathbf{p}^\star_\tau=\operatorname{softmax}(\mathbf{o}^\star_i/\tau)\) and \(\mathbf{p}_\tau=\operatorname{softmax}(\mathbf{o}_i/\tau)\) be the teacher's and the student's full-vocabulary distributions at temperature \(\tau\), and let \(P\) and \(Q\) be the mass each puts on \(S\). Since the state renormalizes the teacher's probabilities on \(S\) (Eq.~\ref{eq:state}), \(p^\star_\tau(v)=P\,s^\star_{i+1}(v)\) for \(v\in S\). Then
\begin{equation}
    \mathcal{D}_k\big(\mathbf{o}^\star_i\,\|\,\mathbf{o}_i\big)
    = \tau^2\Big[\sum_{v\in S} P\,s^\star_{i+1}(v)\log\frac{P\,s^\star_{i+1}(v)}{p_\tau(v)} \;+\; (1-P)\log\frac{1-P}{1-Q}\Big],
\end{equation}
where \(\tau^2\) is the usual distillation scaling~\citep{hinton2015distilling}. Keeping only the terms on \(S\) would push the student to lower the probability of every token outside \(S\), a constant pressure to sharpen its distribution. With the tail, the loss is minimized exactly when the student matches the teacher on \(S\) and puts mass \(1-P\) outside it; and as a KL between coarsened distributions, it is a lower bound on the full-vocabulary KL that does not decrease as \(k\) grows.

\subsection{Adaptation Phase}
\label{app:adaptation}
In the last \(10\%\) of training, each microbatch draws its number of passes \(R\in\{2,3\}\) with equal probability and runs:
\begin{enumerate}[leftmargin=*,itemsep=2pt,topsep=3pt]
    \item \textbf{Pass 1} (no gradient). Every position after the first is fed the bias \(\mathbf{b}\), which carries no information about the context; the first position keeps its learned initial state (\S\ref{sec:architecture}). The logits at every position give a state (Eq.~\ref{eq:state}).
    \item \textbf{Pass 2} (no gradient, only when \(R=3\)). Every position after the first is fed the state that pass 1 predicted at the preceding position, over the whole sequence, yielding refined states.
    \item \textbf{Final pass} (with gradient). Each sequence independently draws, with probability \(p\), a prefix length \(m\sim\mathcal U\{0,\dots,n\}\) (Eq.~\ref{eq:prefix-dropout}). Positions \(1<i\le m\) are fed \(\mathbf{b}\), and all later positions the states of the previous pass. This is the only pass that carries a loss: the language-modeling term of Eq.~\ref{eq:objective} over every position, with the alignment term off (\(\lambda=0\)).
\end{enumerate}
The fed-back states are detached, so no gradient flows between passes, and the no-gradient passes add \(1.5\) forward passes per step on average (Table~\ref{tab:overhead}). The teacher is not needed in this phase.

\FloatBarrier

\section{Experimental Setup}
\label{app:setup}

\subsection{\texorpdfstring{$S_5$}{S5} State Tracking: Setup}
\label{app:s5-setup}

\paragraph{Models and training.}
All models of \S\ref{sec:synthetic} use the OLMo~2 block (\S\ref{app:impl}), slightly adapted to their tiny size and small vocabulary: two layers, width $256$, four attention heads of dimension $64$, a SwiGLU MLP of hidden size $512$, RoPE with $\theta=10{,}000$, and untied input and output embeddings over a synthetic vocabulary of $1{,}047$ tokens (padded to $1{,}152$) that includes the $120$ elements of $S_5$. They are trained in fp32 with AdamW ($\beta=(0.9,0.95)$, $\epsilon=10^{-8}$, weight decay $0.1$ on weight matrices), a peak learning rate of $10^{-3}$ with $200$ warm-up steps and a cosine decay to $0.01\times$ the peak, gradient clipping at $1.0$, and batches of $512$ sequences. Each step draws a fresh batch of a single length, chosen uniformly from the training lengths $N \in \{1,2,3,4,6,8,12,16\}$. The initial state is uniform over $S_5$, and each $a_i$ is drawn uniformly from $12$ fixed generators, which include a transposition and a $5$-cycle and thus generate $S_5$. The input is $\langle \texttt{BOS}, s_0, a_1, \dots, a_N, \texttt{=} \rangle$, and the answer is predicted at the last position.

\paragraph{\modelname{} on \texorpdfstring{$S_5$}{S5}.}
These models differ from the recipe of \S\ref{sec:method} in several respects. The fusion layer is a single linear map, $\hat{\mathbf{h}}^0_i = W_f\,[\mathbf{h}^0_i \,;\, W_e^{\top}\mathbf{s}_i]$ with $W_f \in \mathbb{R}^{d \times 2d}$, in place of Eq.~\ref{eq:fusion}, and the state keeps the top $k=256$ tokens at $\tau=1$. \modelname{}-RNN is trained on the answer loss alone, backpropagating through all positions. The teacher-fed models add the forward KL to the teacher's full output distribution at every position before the answer, with weight $1$, and use no prefix state dropout. The teacher's states are built from its logits, computed in one parallel pass for the Transformer$\times$8 teacher and sequentially, as at inference, for the \modelname{}-RNN teacher. In the last $10\%$ of steps, the teacher-fed models are instead fed their own states, computed sequentially with no gradient through the states, and the losses are unchanged.

\paragraph{Evaluation.}
We report the accuracy of the argmax prediction at the answer position. \modelname{} models are evaluated sequentially, as at inference, each position fed the state predicted at the preceding one; the Transformer answers in one parallel pass. Each length has one evaluation set, shared by all models and seeds and excluded from training: $2{,}000$ sequences, except at $N{=}1$ ($200$) and $N{=}2$ ($1{,}728$), which have only $1{,}440$ and $17{,}280$ distinct sequences. All models are trained with three seeds ($0$, $1$ and $2$), and each teacher-fed \modelname{} model uses the teacher of its own seed, giving three distinct student--teacher pairs.

\subsection{Implementation Details}
\label{app:impl}

\paragraph{Training budgets and teachers.}
The three-scale experiments use training budgets of $5\times$ the Chinchilla token budget~\citep{hoffmann2022training}: 13.4B, 34.9B, and 107.4B tokens for the 135M, 350M, and 1B models, respectively (rounded in Table~\ref{tab:model-configs}). At each scale, we use a pretrained vanilla baseline of the same size as one teacher and a larger pretrained model as the other. The larger teachers for the 135M and 350M models are pretrained vanilla baselines; for the 1B model, we use pretrained OLMo~2-7B~\citep{olmo2024olmo2}. Table~\ref{tab:teachers} lists the teacher sizes and checkpoint training budgets. The compute-matched budgets are given in \S\ref{app:compute-matched}.

\paragraph{Model configurations.}
All three models use the OLMo~2 block~\citep{olmo2024olmo2}: RoPE ($\theta=500{,}000$), RMSNorm with QK-norm, a SwiGLU MLP~\citep{shazeer2020glu} with ratio $8$, a head dimension of $128$, and the dolma2 tokenizer (vocabulary $100{,}278$, padded to $100{,}352$). The 1B model is OLMo~2's 1B configuration unchanged; the 135M and 350M models scale down its width and depth, and tie the input and output embeddings. Table~\ref{tab:model-configs} lists the configurations; model names refer to non-embedding parameter counts.

\begin{table}[ht]
\centering
\small
\caption{Model and training configurations. Parameter counts exclude the RMSNorm vectors; the 1B counts both embedding matrices, the smaller models one tied matrix.}
\label{tab:model-configs}
\begin{tabular}{@{}lrrr@{}}
\toprule
 & 135M & 350M & 1B \\
\midrule
Non-embedding parameters & 134.3M & 348.9M & 1,073.7M \\
Total parameters & 237.0M & 490.2M & 1,484.9M \\
Hidden dimension $d$ & 1024 & 1408 & 2048 \\
Layers & 8 & 11 & 16 \\
Attention heads & 8 & 11 & 16 \\
Embeddings & tied & tied & untied \\
Context length & 1024 & 2048 & 4096 \\
Global batch (sequences) & 1040 & 512 & 512 \\
Tokens per step & 1.06M & 1.05M & 2.10M \\
Training budget ($5\times$ Chinchilla) & 13.4B & 34.9B & 107B \\
Warm-up (steps / tokens) & 2000 / 2.1B & 3000 / 3.1B & 4000 / 8.4B \\
Peak learning rate & $6\times10^{-4}$ & $5\times10^{-4}$ & $4\times10^{-4}$ \\
\bottomrule
\end{tabular}
\end{table}

\paragraph{Optimization.}
We use AdamW with $\beta=(0.9, 0.95)$, $\epsilon=10^{-8}$, weight decay $0.1$ (not applied to the embeddings), gradient clipping at a norm of $1.0$, and a z-loss with weight $10^{-5}$, all as in OLMo~2. Training runs in bf16 mixed precision with the cross-entropy loss computed in fp32. The state input, the teacher's top-$k$ distribution, is quantized to 8 bits. Within each model size, \modelname{} and its vanilla control share the configuration, the random seed and therefore the exact data order.

\paragraph{\modelname{} hyperparameters.}
The three-scale experiments use the same \modelname{} settings at every scale. The state keeps the top \(k=1024\) tokens at temperature \(\tau=1.5\) (Eq.~\ref{eq:state}); the alignment term uses the same \(k\) and \(\tau\), with weight \(\lambda=1.5\) (Eq.~\ref{eq:objective}). Prefix state dropout is applied to each training sequence with probability \(p=0.5\) (\S\ref{app:prefix-dropout}). The adaptation phase covers the last \(10\%\) of training, coinciding with the learning-rate anneal, and draws one or two no-gradient passes per step with equal probability (\S\ref{app:adaptation}). The comparison of \S\ref{sec:jacobi} uses its own settings (\S\ref{app:jacobi-setup}).

\paragraph{Pretraining data.}
The training corpus is sampled from the OLMo~2 stage-1 pretraining mix: DCLM-baseline, StarCoder, peS2o, Proof-Pile-2 and OLMo-Mix, tokenized with the dolma2 tokenizer. Our stratified versions sample whole files per source domain so that the source mixture is preserved: approximately $93.8\%$ DCLM, $2.6\%$ StarCoder, $2.0\%$ peS2o, $1.3\%$ Proof-Pile-2 and $0.4\%$ OLMo-Mix. We use a 400B-token selection for the 350M and 1B runs and a 40B-token selection for the 135M runs; every run consumes its budget from a single pass over a random order of its selection.

\paragraph{Learning-rate schedule.}
All models follow the WSD protocol of \citet{wen2024wsd}. The learning rate warms up linearly from zero over the steps listed in Table~\ref{tab:model-configs}, is then held constant at the peak value for the first $90\%$ of the training budget ($4.5\times$ Chinchilla), and is annealed over the last $10\%$ ($0.5\times$ Chinchilla), branching from the trunk checkpoint at $90\%$. The anneal uses the inverse-proportional decay shape of \citet{wen2024wsd}, in which the reciprocal of the learning rate rises linearly, and ends at $0.1\times$ the peak value. All results reported in \S\ref{sec:experiments} are after an identical annealing over the last 10\% of training to an LR of $0.1\times$ the starting value.

\subsection{Feedback-Transformer Comparison: Setup}
\label{app:jacobi-setup}

\paragraph{Setup.}
All models of \S\ref{sec:jacobi} are trained with the T\textsuperscript{2}MLR authors' training code on the SmolLM2-135M backbone~\citep{allal2025smollm2smolgoesbig}: 30 layers, width 576, 9 attention heads with 3 key-value heads, tied embeddings, 135M parameters including embeddings, and the SmolLM2 tokenizer of 49{,}152 tokens. The data is the 10B-token FineWeb-Edu sample (\texttt{sample-10BT}; \citealp{penedo2024finewebdatasetsdecantingweb}) packed into sequences of 2{,}048 tokens; one epoch is 19{,}073 steps of 256 sequences. The training configuration is identical for all models: AdamW with $\beta=(0.9,0.98)$ and weight decay $0.01$, a peak learning rate of $5\times10^{-4}$ with a warm-up of 954 steps ($5\%$), a global batch of 256 sequences ($0.5$M tokens per step), gradient clipping at $1.0$, bf16 precision, and a WSD schedule with a linear decay to $0.1\times$ the peak over the last $10\%$ of steps; for \modelname{} that last $10\%$ is the adaptation phase of \S\ref{sec:training}. T\textsuperscript{2}MLR runs in the authors' configuration, T\textsuperscript{2}MLR(13,18) with 16 forward and 4 backward refinement passes, and is evaluated with the exact sequential recurrence, as in their paper. \modelname{} uses $k=1024$ and is trained without the KL objective, so that the measured gain comes from state propagation and not from the distillation signal; its teacher is the last checkpoint of the vanilla baseline model (the same model, trained on 23B tokens). The compute-matched budget is T\textsuperscript{2}MLR's 19{,}073 steps at $2.28\times$ the FLOPs of a vanilla step per token; at $1.36\times$ per token in the trunk and $1.87\times$ in the adaptation phase, \modelname{} reaches it after 16.2B tokens and the vanilla model after 23.0B. The checkpoint the authors published was trained for 19{,}294 steps against our 19{,}073 ($1.2\%$ more tokens), with a cosine decay to $0.001\times$ the peak. Parameter counts are 138.2M for \modelname{}, 136.2M for T\textsuperscript{2}MLR, 135.2M for the Multi-pass Transformer and 134.5M for the vanilla model. Perplexity is measured in fp32 on 127K held-out FineWeb-Edu tokens; the arithmetic and LM-eval protocols are those of \S\ref{app:benchmarks}.

\paragraph{Multi-pass Transformer.}
We take from the paper~\citep{wang2026fullbandwidth} its feedback path, a gated linear unit that fuses the previous top-layer state (the value) with the current token embedding (the gate); its pass schedule, \(75\%\) of the batches with a single pass, \(22\%\) with two and \(3\%\) with three, which is the schedule the authors report for their larger runs; its prefix mixin, which leaves a random prefix of each sequence on plain embeddings in the extra passes, so that training matches the prompt-then-generation structure of inference; and its stabilizers, noise on the carried state and normalization of the fused input (the third, tying the input and output embeddings, the shared backbone already does). We could not take their optimizer, since all models here share one and changing it for a single model would confound the comparison, nor their depth scaling of the residual branches, which alters the backbone shared by all models here and which we replace by carrying the normalized top-layer state. We use the shared 135M, 10B-token setup instead of their 1B model and 400B-token dataset. As with T\textsuperscript{2}MLR, we report the model in its recurrent mode.
\subsection{Evaluation Benchmarks}
\label{app:benchmarks}

\paragraph{Downstream suite.}
The twenty suites are run with OLMES~\citep{gu2024olmes} and its task configurations (Table~\ref{tab:olmes-suites}); \modelname{} scores every prompt and answer with its own states. The window is each model's trained context (1{,}024 / 2{,}048 / 4{,}096 tokens at 135M / 350M / 1B; 2{,}048 for the SmolLM2 models), and at 135M, generation is capped at 256 tokens, the same for every 135M model, so its generative numbers compare within the size, not across sizes. Cloze scoring appends each answer option to the question and scores it by its length-normalized log-likelihood; the prediction is the highest-scoring option. Bits per byte is $-\log_2 P(\text{gold answer})$ divided by the answer's UTF-8 bytes. The main table keeps a suite if, at 1B, at least one model is five points above chance (multiple choice) or above 5\% (generative); bits-per-byte suites always enter. This leaves out the letter-format multiple choice (MMLU, AGI-Eval), where every model is at chance, and generative math and code (GSM8K, MATH, HumanEval, MBPP), where every model is under 3\%. BoolQ is left out as well: a yes/no cloze whose majority class is 62\%, which several models fall below and which moves by ten points across training seeds. Every excluded suite is reported in Tables~\ref{tab:olmes-135m}--\ref{tab:olmes-jacobi}.

\begin{table}[ht]
\centering
\footnotesize
\setlength{\tabcolsep}{3pt}
\renewcommand{\arraystretch}{1.15}
\caption{The downstream suite. Items: scored instances (per task for the core tasks); Shots: in-context examples; Group: the Table~\ref{tab:lm} column, or the reason a suite is reported only in the appendix. Task sources are cited below the table.}
\label{tab:olmes-suites}
\begin{tabular}{@{}>{\raggedright\arraybackslash}p{3.3cm}>{\raggedright\arraybackslash}p{3.0cm}>{\raggedright\arraybackslash}p{2.25cm}>{\raggedright\arraybackslash}p{1.4cm}>{\raggedright\arraybackslash}p{1.45cm}>{\raggedright\arraybackslash}p{1.4cm}@{}}
\toprule
Suite & Tests & Items & Shots & Scoring & Group \\
\midrule
Core 8: ARC-Easy, ARC-Challenge, CommonsenseQA, HellaSwag, OpenBookQA, PIQA, SocialIQA, WinoGrande & Science and commonsense questions & 500--1{,}267 & 5 & cloze acc. & MC \\
MMLU & 57 academic subjects & 14{,}042 & 5 & cloze acc. & MC \\
Basic skills (OLMES) & Arithmetic, string operations, pattern continuation, coding, logical reasoning, common knowledge & 5{,}967 & 5 & cloze acc. & MC \\
Generative QA: CoQA, SQuAD, Jeopardy, Natural Questions, DROP & Conversational, extractive and closed-book QA & 7{,}983 / 1{,}000 / 2{,}116 / 1{,}000 / 1{,}000 & 5 (CoQA 0) & F1 & Gen. \\
TriviaQA & Closed-book trivia & 7{,}993 & 5 & F1 & Gen. \\
BBH & 27 reasoning tasks with chain of thought & 6{,}511 & 3 & exact match & Gen. \\
ARC, MMLU, basic skills; MATH; HumanEval; MBPP & The gold answer's likelihood & 3{,}548 / 14{,}042 / 5{,}967 / 5{,}000 / 164 / 500 & 5 / 5 / 5 / 4 / 3 / 3 & bits per byte & RC \\
\midrule
BoolQ & Yes/no questions on a passage & 1{,}000 & 5 & cloze acc. & coin flip \\
MMLU, letter format & As above, scored by the option letter & 14{,}042 & 5 & letter acc. & at chance \\
AGI-Eval English & Exam questions, letter format & 2{,}646 & 1 & letter acc. & at chance \\
GSM8K & Grade-school math, chain of thought & 1{,}319 & 5 and 8 & exact match & under 3\% \\
MATH-500, Minerva MATH & Competition math & 500, 5{,}000 & 4 & exact match & under 3\% \\
HumanEval, MBPP & Program synthesis, 20 samples at $T{=}0.8$ & 164, 500 & 3 & pass@1 & under 3\% \\
\bottomrule
\end{tabular}
\par\smallskip
\begin{minipage}{\linewidth}\footnotesize
Sources: ARC~\citep{clark2018think}, CommonsenseQA~\citep{talmor2019commonsenseqa}, HellaSwag~\citep{zellers2019hellaswag}, OpenBookQA~\citep{mihaylov2018can}, PIQA~\citep{bisk2020piqa}, SocialIQA~\citep{sap2019socialiqa}, WinoGrande~\citep{sakaguchi2021winogrande}, MMLU~\citep{hendrycks2021mmlu}, CoQA~\citep{reddy2019coqa}, SQuAD~\citep{rajpurkar2016squad}, Natural Questions~\citep{kwiatkowski2019natural}, DROP~\citep{dua2019drop}, TriviaQA~\citep{joshi2017triviaqa}, BBH~\citep{suzgun2023bbh}, MATH~\citep{hendrycks2021math}, HumanEval~\citep{chen2021codex}, MBPP~\citep{austin2021mbpp}, BoolQ~\citep{clark2019boolq}, AGI-Eval~\citep{zhong2023agieval}, GSM8K~\citep{cobbe2021gsm8k}.
\end{minipage}
\end{table}

\paragraph{Arithmetic.}
Expressions are generated by drawing $n$ integers uniformly from $0$ to $9$ and $n-1$ operators uniformly from $\{+,-\}$, and are evaluated left to right; duplicates are discarded, which is why the $n=2$ bin holds all 200 possible expressions. The answer is an integer, possibly negative. Every prompt consists of the same three worked examples (3-shot), followed by the query, one expression per line, and the scored continuation is the answer with its leading space:
\begin{quote}
\texttt{4 - 6 = -2}\\
\texttt{0 + 5 + 7 - 1 - 3 = 8}\\
\texttt{3 - 3 - 1 = -1}\\
\texttt{2 + 0 - 6 + 7 + 5 =}
\end{quote}
with answer \texttt{8}. Bits per byte is $-\log_2 P(\text{answer})$ summed over the set and divided by the total number of UTF-8 bytes of the scored answers. The exact-match accuracy of the teacher-forced argmax is recorded alongside it (Table~\ref{tab:arith-acc-by-length}): it is substantial only at two operands and a few percent from three operands on, for every model, which is why Table~\ref{tab:lm} reports bits per byte.

\FloatBarrier

\section{Additional Results}
\label{app:additional_results}

\subsection{Transformer Training Budget on \texorpdfstring{$S_5$}{S5}}
\label{app:transformer-budget}

\begin{figure}[!htbp]
    \centering
    \includegraphics[width=0.8\linewidth]{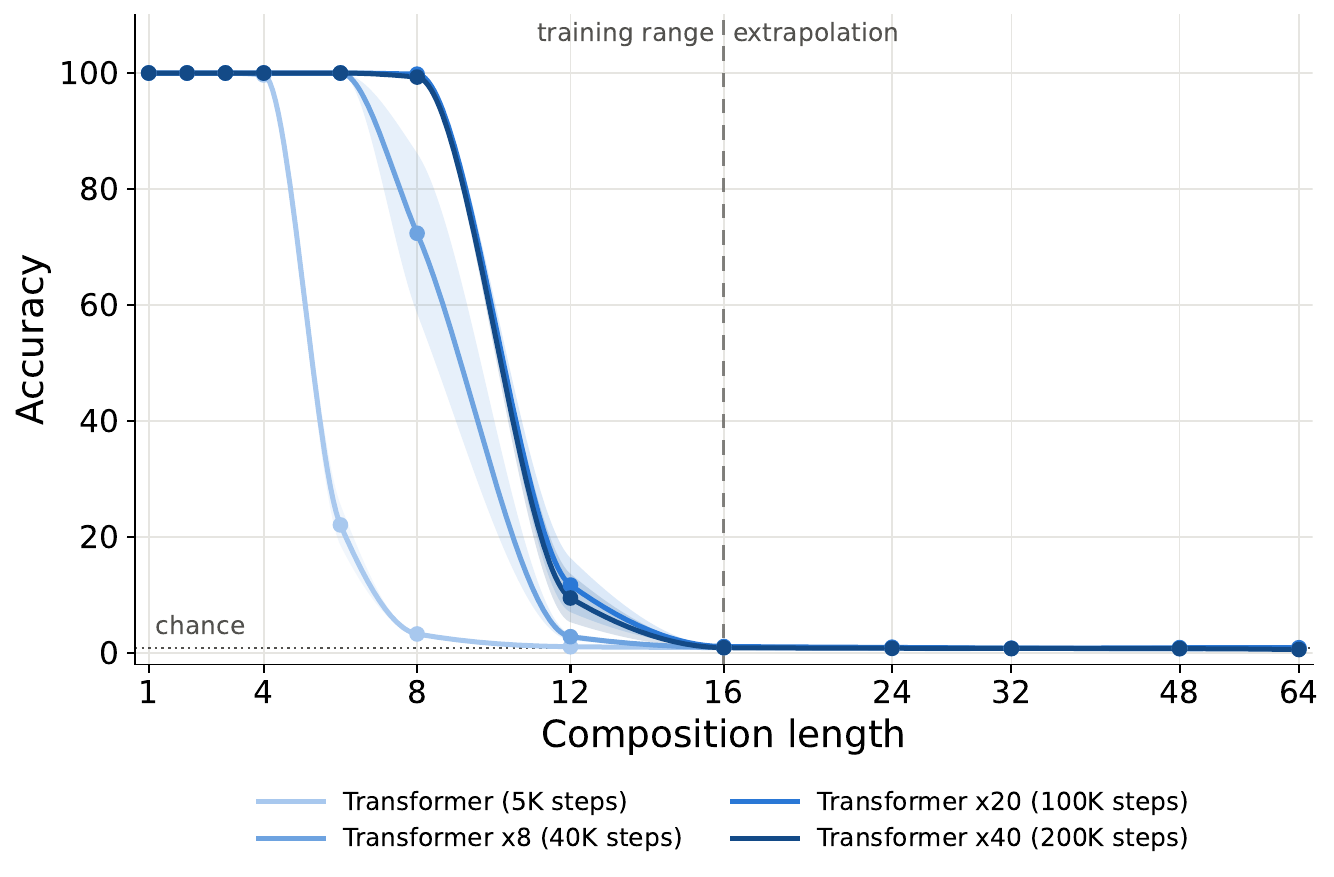}
    \caption{The same two-layer Transformer on $S_5$ at four training budgets, darker lines for longer training. Bands show standard error over three seeds. Training beyond 100K steps ($20\times$) brings no further gain.}
    \label{fig:s5-budget}
\end{figure}

\S\ref{sec:synthetic} attributes the two-layer Transformer's ceiling on $S_5$ to depth. Every $S_5$ model has $d_{\text{model}}=256$ and trains with batch size $512$ for $5{,}000$ steps unless stated otherwise. A competing explanation is that the model is simply undertrained, so we trained the same baseline at four budgets, all else held fixed: the token-matched $5{,}000$ steps, $8\times$ ($40{,}000$), $20\times$ ($100{,}000$) and $40\times$ ($200{,}000$). Figure~\ref{fig:s5-budget} reports all four, with three seeds each, under the protocol of \S\ref{sec:synthetic}.

The model gains nothing after $20\times$, where the $20\times$ and $40\times$ curves coincide, and it remains at $\approx10\%$ at $N{=}12$ and at chance at $N{=}16$ and every longer length, in every seed of every budget.

\subsection{Perplexity by Source and Evaluation Uncertainty}
\label{app:res-ppl}

Table~\ref{tab:ppl-by-source} reports perplexity on every source of the OLMo~2 perplexity evaluation set: the full C4 validation subset and $10\%$ of each other source. \modelname{} is ahead of the token-matched, compute-matched and distillation baselines on all eleven sources at every size. Table~\ref{tab:ppl-bootstrap} gives, for the paper's training seed, a paired bootstrap interval over the evaluation sequences for every perplexity gap of Table~\ref{tab:lm}; it measures how much a gap depends on the held-out text at a fixed training run, not seed variance.

\begin{table}[!htbp]
\centering
\small
\setlength{\tabcolsep}{5pt}
\renewcommand{\arraystretch}{1.1}
\caption{Perplexity by source on the OLMo~2 perplexity evaluation set: the full C4 validation slice (971 sequences of 1{,}024 tokens) and $10\%$ of each other source (at least 50 sequences). Columns as in Table~\ref{tab:lm}; lower is better, bold is the best in a row.}
\label{tab:ppl-by-source}
\begin{tabular}{@{}lrrrrrr@{}}
\toprule
Source & \modelname{} & w/o states & Tr.\ token & Tr.\ compute & Distillation & Seqs \\
\midrule
\multicolumn{7}{@{}l}{\textit{135M}} \\
C4 & \textbf{30.65} & 32.44 & 32.43 & 31.04 & 31.77 & 971 \\
Common Crawl & \textbf{31.55} & 33.44 & 33.55 & 32.01 & 32.63 & 50 \\
Books & \textbf{33.58} & 35.93 & 35.90 & 33.85 & 34.86 & 50 \\
Wikipedia & \textbf{19.64} & 20.70 & 20.78 & 19.87 & 20.39 & 50 \\
peS2o & \textbf{17.85} & 18.93 & 19.00 & 18.12 & 18.56 & 51 \\
S2ORC & \textbf{38.88} & 40.91 & 41.50 & 39.69 & 40.34 & 97 \\
Reddit & \textbf{33.89} & 35.83 & 35.74 & 34.37 & 34.88 & 50 \\
Stack (code) & \textbf{7.10} & 7.45 & 7.59 & 7.20 & 7.38 & 50 \\
Pile & \textbf{18.20} & 19.29 & 19.30 & 18.52 & 18.87 & 66 \\
ICE & \textbf{30.52} & 32.98 & 33.92 & 31.23 & 32.28 & 89 \\
WikiText-103 & \textbf{24.69} & 26.72 & 26.88 & 25.23 & 26.30 & 50 \\
\midrule
\multicolumn{7}{@{}l}{\textit{350M}} \\
C4 & \textbf{23.82} & 24.96 & 25.15 & 24.32 & 24.52 & 971 \\
Common Crawl & \textbf{24.70} & 25.90 & 26.02 & 25.06 & 25.32 & 50 \\
Books & \textbf{24.29} & 25.69 & 25.94 & 24.76 & 25.12 & 50 \\
Wikipedia & \textbf{15.85} & 16.52 & 16.75 & 16.11 & 16.29 & 50 \\
peS2o & \textbf{12.48} & 12.99 & 13.04 & 12.67 & 12.77 & 51 \\
S2ORC & \textbf{27.44} & 28.68 & 29.32 & 28.46 & 28.15 & 97 \\
Reddit & \textbf{26.79} & 28.17 & 28.30 & 27.49 & 27.65 & 50 \\
Stack (code) & \textbf{5.94} & 6.13 & 6.25 & 6.04 & 6.04 & 50 \\
Pile & \textbf{14.07} & 14.77 & 14.97 & 14.49 & 14.64 & 66 \\
ICE & \textbf{24.51} & 26.00 & 26.73 & 25.61 & 26.44 & 89 \\
WikiText-103 & \textbf{18.43} & 19.56 & 19.59 & 18.86 & 19.07 & 50 \\
\midrule
\multicolumn{7}{@{}l}{\textit{1B}} \\
C4 & \textbf{18.81} & 19.62 & 19.79 & 19.17 & 19.40 & 971 \\
Common Crawl & \textbf{19.62} & 20.37 & 20.59 & 19.96 & 20.16 & 50 \\
Books & \textbf{17.82} & 18.71 & 19.05 & 18.27 & 18.44 & 50 \\
Wikipedia & \textbf{12.31} & 12.73 & 12.88 & 12.46 & 12.68 & 50 \\
peS2o & \textbf{10.07} & 10.39 & 10.53 & 10.26 & 10.33 & 51 \\
S2ORC & \textbf{22.38} & 23.05 & 23.69 & 23.12 & 23.63 & 97 \\
Reddit & \textbf{21.74} & 22.63 & 22.78 & 22.22 & 22.34 & 50 \\
Stack (code) & \textbf{4.85} & 4.97 & 5.03 & 4.89 & 4.93 & 50 \\
Pile & \textbf{11.28} & 11.71 & 11.87 & 11.50 & 11.61 & 66 \\
ICE & \textbf{19.34} & 20.23 & 20.43 & 19.91 & 19.96 & 89 \\
WikiText-103 & \textbf{13.79} & 14.36 & 14.45 & 13.96 & 14.17 & 50 \\
\bottomrule
\end{tabular}
\end{table}

\begin{table}[!htbp]
\centering
\small
\setlength{\tabcolsep}{6pt}
\renewcommand{\arraystretch}{1.1}
\caption{Evaluation-set uncertainty of the perplexity gaps of Table~\ref{tab:lm}, for the paper's training seed: the gap of \modelname{} to each Transformer baseline with a paired bootstrap $95\%$ interval over the evaluation sequences (both models scored on the same text; 5{,}000 resamples). This measures how much a gap depends on the held-out text, at a fixed training run; it is not seed variance, which Table~\ref{tab:olmes-seeds} reports for 135M. Negative is in favor of \modelname{}.}
\label{tab:ppl-bootstrap}
\begin{tabular}{@{}lrrr@{}}
\toprule
$\Delta$ PPL & 135M & 350M & 1B \\
\midrule
vs.\ Transformer, token-match & -1.77 [-1.85, -1.70] & -1.33 [-1.39, -1.28] & -0.99 [-1.05, -0.93] \\
vs.\ Transformer, compute-match & -0.38 [-0.44, -0.33] & -0.50 [-0.55, -0.46] & -0.36 [-0.40, -0.32] \\
vs.\ Distillation & -1.11 [-1.17, -1.06] & -0.70 [-0.74, -0.66] & -0.59 [-0.63, -0.55] \\
\bottomrule
\end{tabular}
\end{table}

\subsection{Downstream Results per Suite}
\label{app:res-downstream}

Tables~\ref{tab:olmes-135m}--\ref{tab:olmes-1b} give every OLMES suite for the rows of Tables~\ref{tab:lm} and~\ref{tab:teachers}, including the suites left out of the main table.

\begin{table}[!htbp]
\centering
\footnotesize
\setlength{\tabcolsep}{2.2pt}
\caption{OLMES results at 135M for the rows of Tables~\ref{tab:lm} and~\ref{tab:teachers}: the suites of the three Table~\ref{tab:lm} groups, then the suites left out of it (at chance or floor for every model, and BoolQ). Accuracy \% for the multiple-choice, generative and excluded rows; bits per byte (lower is better) for the BPB rows; suites as in Table~\ref{tab:olmes-suites}. Gray: the 2$\times$-token teacher, outside the bold rule. Bold: best in the row.}
\label{tab:olmes-135m}
\begin{tabular}{@{}lcccccc@{}}
\toprule
Suite & \begin{tabular}[c]{@{}c@{}}\modelname{}\\{\scriptsize 135M@29.6B}\end{tabular} & \begin{tabular}[c]{@{}c@{}}w/o states\\{\scriptsize 135M@29.6B}\end{tabular} & \begin{tabular}[c]{@{}c@{}}Tr.\ token\\{\scriptsize 13.4B}\end{tabular} & \begin{tabular}[c]{@{}c@{}}Tr.\ compute\\{\scriptsize 19.3B}\end{tabular} & \begin{tabular}[c]{@{}c@{}}Distillation\\{\scriptsize 13.4B}\end{tabular} & \begin{tabular}[c]{@{}c@{}}2$\times$ tokens\\{\scriptsize 29.6B}\end{tabular} \\
\midrule
Core 8 (cloze) & \textbf{44.0} & 43.3 & 42.5 & 43.9 & 43.9 & \textcolor{black!65}{44.7} \\
MMLU (cloze) & \textbf{28.1} & 27.9 & 27.3 & 28.0 & 27.6 & \textcolor{black!65}{28.2} \\
Basic skills (cloze) & 46.0 & 44.8 & 44.6 & \textbf{46.4} & 45.8 & \textcolor{black!65}{47.2} \\
Generative QA & 13.4 & 11.3 & 11.4 & \textbf{13.8} & 10.7 & \textcolor{black!65}{15.5} \\
TriviaQA & 13.6 & 13.3 & 13.8 & \textbf{14.8} & 14.3 & \textcolor{black!65}{14.8} \\
BBH (CoT) & \textbf{19.5} & 11.3 & 15.6 & 13.8 & 12.9 & \textcolor{black!65}{14.7} \\
ARC BPB & 1.039 & 1.076 & 1.082 & \textbf{1.032} & 1.039 & \textcolor{black!65}{0.990} \\
MMLU BPB & \textbf{1.298} & 1.336 & 1.347 & 1.314 & 1.313 & \textcolor{black!65}{1.276} \\
Basic skills BPB & 1.338 & 1.367 & 1.377 & \textbf{1.327} & 1.340 & \textcolor{black!65}{1.308} \\
MATH BPB & \textbf{0.998} & 1.024 & 1.022 & 1.000 & 1.013 & \textcolor{black!65}{0.976} \\
HumanEval BPB & \textbf{0.842} & 0.866 & 0.853 & \textbf{0.842} & 0.871 & \textcolor{black!65}{0.843} \\
MBPP BPB & \textbf{0.951} & 0.965 & 0.976 & 0.967 & 0.960 & \textcolor{black!65}{0.947} \\
\midrule
\multicolumn{7}{l}{\emph{Not in Table~\ref{tab:lm}}} \\
BoolQ (cloze) & 63.8 & 63.8 & 59.9 & 58.8 & \textbf{63.9} & \textcolor{black!65}{63.3} \\
MMLU (letter) & 25.3 & \textbf{26.1} & \textbf{26.1} & \textbf{26.1} & 25.7 & \textcolor{black!65}{24.9} \\
AGI-Eval (letter) & 21.6 & 21.9 & 22.3 & 21.6 & \textbf{23.6} & \textcolor{black!65}{23.0} \\
GSM8K (5-shot) & \textbf{2.3} & \textbf{2.3} & 1.4 & 1.7 & 1.8 & \textcolor{black!65}{2.3} \\
GSM8K (8-shot) & 2.2 & 2.0 & \textbf{2.7} & 1.9 & 2.5 & \textcolor{black!65}{2.6} \\
MATH-500 & \textbf{0.2} & 0.0 & \textbf{0.2} & 0.0 & \textbf{0.2} & \textcolor{black!65}{0.2} \\
Minerva MATH & \textbf{0.2} & 0.0 & 0.0 & 0.0 & 0.1 & \textcolor{black!65}{0.2} \\
HumanEval pass@1 & \textbf{0.2} & 0.1 & 0.1 & 0.1 & 0.0 & \textcolor{black!65}{0.2} \\
MBPP pass@1 & \textbf{0.0} & \textbf{0.0} & \textbf{0.0} & \textbf{0.0} & \textbf{0.0} & \textcolor{black!65}{0.0} \\
\bottomrule
\end{tabular}
\end{table}

\begin{table}[!htbp]
\centering
\footnotesize
\setlength{\tabcolsep}{2.2pt}
\caption{OLMES results at 350M for the rows of Tables~\ref{tab:lm} and~\ref{tab:teachers}: the suites of the three Table~\ref{tab:lm} groups, then the suites left out of it (at chance or floor for every model, and BoolQ). Accuracy \% for the multiple-choice, generative and excluded rows; bits per byte (lower is better) for the BPB rows; suites as in Table~\ref{tab:olmes-suites}. Gray: the 2$\times$-token teacher, outside the bold rule. Bold: best in the row.}
\label{tab:olmes-350m}
\begin{tabular}{@{}lcccccc@{}}
\toprule
Suite & \begin{tabular}[c]{@{}c@{}}\modelname{}\\{\scriptsize 350M@77.8B}\end{tabular} & \begin{tabular}[c]{@{}c@{}}w/o states\\{\scriptsize 350M@77.8B}\end{tabular} & \begin{tabular}[c]{@{}c@{}}Tr.\ token\\{\scriptsize 34.9B}\end{tabular} & \begin{tabular}[c]{@{}c@{}}Tr.\ compute\\{\scriptsize 49.9B}\end{tabular} & \begin{tabular}[c]{@{}c@{}}Distillation\\{\scriptsize 34.9B}\end{tabular} & \begin{tabular}[c]{@{}c@{}}2$\times$ tokens\\{\scriptsize 77.8B}\end{tabular} \\
\midrule
Core 8 (cloze) & \textbf{51.2} & 49.8 & 48.9 & 50.8 & 50.4 & \textcolor{black!65}{51.7} \\
MMLU (cloze) & 30.2 & 30.1 & 30.3 & \textbf{31.0} & 30.6 & \textcolor{black!65}{31.1} \\
Basic skills (cloze) & \textbf{53.8} & 52.7 & 50.3 & 50.4 & 52.4 & \textcolor{black!65}{52.6} \\
Generative QA & \textbf{27.3} & 24.0 & 21.5 & 23.1 & 24.8 & \textcolor{black!65}{26.6} \\
TriviaQA & \textbf{21.9} & 20.6 & 20.1 & 21.7 & 21.4 & \textcolor{black!65}{23.6} \\
BBH (CoT) & 20.3 & 18.4 & \textbf{21.3} & \textbf{21.3} & 20.1 & \textcolor{black!65}{21.9} \\
ARC BPB & \textbf{0.872} & 0.893 & 0.905 & 0.893 & 0.875 & \textcolor{black!65}{0.865} \\
MMLU BPB & \textbf{1.139} & 1.164 & 1.173 & 1.154 & 1.155 & \textcolor{black!65}{1.129} \\
Basic skills BPB & \textbf{1.099} & 1.136 & 1.169 & 1.153 & 1.129 & \textcolor{black!65}{1.153} \\
MATH BPB & \textbf{0.859} & 0.878 & 0.886 & 0.865 & 0.872 & \textcolor{black!65}{0.849} \\
HumanEval BPB & 0.735 & 0.758 & 0.751 & 0.750 & \textbf{0.729} & \textcolor{black!65}{0.694} \\
MBPP BPB & \textbf{0.825} & 0.840 & 0.851 & 0.839 & 0.832 & \textcolor{black!65}{0.815} \\
\midrule
\multicolumn{7}{l}{\emph{Not in Table~\ref{tab:lm}}} \\
BoolQ (cloze) & \textbf{63.7} & 63.6 & 61.7 & 55.9 & 46.2 & \textcolor{black!65}{58.1} \\
MMLU (letter) & 24.3 & 24.3 & 25.0 & \textbf{26.9} & 25.0 & \textcolor{black!65}{25.9} \\
AGI-Eval (letter) & 22.7 & 22.2 & \textbf{23.0} & 22.2 & 21.4 & \textcolor{black!65}{23.8} \\
GSM8K (5-shot) & 1.4 & 1.4 & \textbf{2.0} & \textbf{2.0} & 1.4 & \textcolor{black!65}{2.7} \\
GSM8K (8-shot) & 2.3 & 1.9 & \textbf{2.4} & 1.8 & 1.7 & \textcolor{black!65}{3.0} \\
MATH-500 & 1.0 & \textbf{1.2} & 0.0 & 0.2 & 0.4 & \textcolor{black!65}{1.0} \\
Minerva MATH & \textbf{0.9} & 0.6 & 0.2 & 0.2 & 0.5 & \textcolor{black!65}{0.8} \\
HumanEval pass@1 & 0.2 & 0.2 & 0.3 & \textbf{0.4} & 0.2 & \textcolor{black!65}{0.5} \\
MBPP pass@1 & 0.1 & \textbf{0.2} & 0.1 & \textbf{0.2} & 0.1 & \textcolor{black!65}{0.3} \\
\bottomrule
\end{tabular}
\end{table}

\begin{table}[!htbp]
\centering
\footnotesize
\setlength{\tabcolsep}{2.2pt}
\caption{OLMES results at 1B for the rows of Tables~\ref{tab:lm} and~\ref{tab:teachers}: the suites of the three Table~\ref{tab:lm} groups, then the suites left out of it (at chance or floor for every model, and BoolQ). Accuracy \% for the multiple-choice, generative and excluded rows; bits per byte (lower is better) for the BPB rows; suites as in Table~\ref{tab:olmes-suites}. Gray: the 2$\times$-token teacher, outside the bold rule. Bold: best in the row.}
\label{tab:olmes-1b}
\begin{tabular}{@{}lcccccccc@{}}
\toprule
Suite & \begin{tabular}[c]{@{}c@{}}\modelname{}\\{\scriptsize 1B@222B}\end{tabular} & \begin{tabular}[c]{@{}c@{}}\modelname{}\\{\scriptsize 1B@4001B}\end{tabular} & \begin{tabular}[c]{@{}c@{}}w/o states\\{\scriptsize 1B@222B}\end{tabular} & \begin{tabular}[c]{@{}c@{}}w/o states\\{\scriptsize 1B@4001B}\end{tabular} & \begin{tabular}[c]{@{}c@{}}Tr.\ token\\{\scriptsize 107.4B}\end{tabular} & \begin{tabular}[c]{@{}c@{}}Tr.\ compute\\{\scriptsize 152.4B}\end{tabular} & \begin{tabular}[c]{@{}c@{}}Distillation\\{\scriptsize 107.4B}\end{tabular} & \begin{tabular}[c]{@{}c@{}}2$\times$ tokens\\{\scriptsize 222B}\end{tabular} \\
\midrule
Core 8 (cloze) & \textbf{59.4} & 58.3 & 58.3 & 56.7 & 58.0 & 59.0 & 58.5 & \textcolor{black!65}{60.2} \\
MMLU (cloze) & \textbf{35.4} & 35.1 & 34.7 & 34.3 & 34.4 & 34.9 & 35.0 & \textcolor{black!65}{35.6} \\
Basic skills (cloze) & \textbf{62.6} & 62.2 & 60.6 & 61.5 & 59.3 & 60.9 & 62.5 & \textcolor{black!65}{63.0} \\
Generative QA & \textbf{42.2} & 42.1 & 39.3 & 39.5 & 38.6 & 40.5 & 41.4 & \textcolor{black!65}{43.5} \\
TriviaQA & 37.3 & 37.6 & 36.1 & 35.7 & 34.9 & \textbf{37.9} & 36.7 & \textcolor{black!65}{41.2} \\
BBH (CoT) & 25.6 & \textbf{27.8} & 23.2 & 26.2 & 25.2 & 26.0 & 26.4 & \textcolor{black!65}{27.4} \\
ARC BPB & 0.753 & \textbf{0.752} & 0.768 & 0.774 & 0.770 & 0.771 & 0.759 & \textcolor{black!65}{0.744} \\
MMLU BPB & \textbf{0.991} & \textbf{0.991} & 1.008 & 1.011 & 1.023 & 1.004 & 1.003 & \textcolor{black!65}{0.979} \\
Basic skills BPB & \textbf{0.845} & 0.866 & 0.885 & 0.874 & 0.955 & 0.919 & 0.884 & \textcolor{black!65}{0.858} \\
MATH BPB & 0.734 & \textbf{0.731} & 0.750 & 0.748 & 0.762 & 0.742 & 0.740 & \textcolor{black!65}{0.726} \\
HumanEval BPB & 0.596 & 0.579 & 0.611 & \textbf{0.577} & 0.599 & 0.589 & 0.584 & \textcolor{black!65}{0.565} \\
MBPP BPB & \textbf{0.711} & 0.716 & 0.724 & 0.733 & 0.740 & 0.728 & 0.717 & \textcolor{black!65}{0.711} \\
\midrule
\multicolumn{9}{l}{\emph{Not in Table~\ref{tab:lm}}} \\
BoolQ (cloze) & 64.9 & 60.4 & 65.7 & 53.2 & 63.8 & 64.2 & \textbf{66.8} & \textcolor{black!65}{55.4} \\
MMLU (letter) & 24.8 & 23.3 & 25.0 & 23.9 & 26.4 & \textbf{27.1} & 25.0 & \textcolor{black!65}{28.9} \\
AGI-Eval (letter) & 21.7 & 21.7 & 20.6 & 20.7 & \textbf{24.6} & 21.7 & 21.4 & \textcolor{black!65}{23.1} \\
GSM8K (5-shot) & 2.0 & 2.6 & 2.0 & 1.9 & 2.9 & 2.7 & \textbf{3.0} & \textcolor{black!65}{2.4} \\
GSM8K (8-shot) & \textbf{2.5} & 2.3 & 2.4 & 2.4 & 2.0 & 2.3 & 2.0 & \textcolor{black!65}{1.8} \\
MATH-500 & \textbf{2.6} & 1.6 & 0.4 & 0.0 & 1.6 & 0.6 & 1.2 & \textcolor{black!65}{0.4} \\
Minerva MATH & \textbf{1.2} & 0.7 & 0.5 & 0.3 & 1.0 & 1.0 & 0.7 & \textcolor{black!65}{0.9} \\
HumanEval pass@1 & \textbf{2.7} & 2.5 & 2.3 & 1.9 & 1.7 & 2.1 & 2.0 & \textcolor{black!65}{2.3} \\
MBPP pass@1 & \textbf{1.0} & 0.8 & 0.7 & 0.5 & 0.4 & 0.7 & \textbf{1.0} & \textcolor{black!65}{1.0} \\
\bottomrule
\end{tabular}
\end{table}

\subsection{Seed Variance at 135M}
\label{app:res-seeds}

Table~\ref{tab:olmes-seeds} gives the 135M rows of Table~\ref{tab:lm} over three training seeds.

\begin{table}[!htbp]
\centering
\small
\setlength{\tabcolsep}{3.4pt}
\caption{The 135M rows of Table~\ref{tab:lm} over three training seeds (6198 / 6199 / 6200): mean $\pm$ sd of each column. We regard a difference between two models as beyond seed noise when their mean $\pm$ sd ranges do not overlap. Pattern: the basic-skills pattern-continuation task (534 questions).}
\label{tab:olmes-seeds}
\resizebox{\linewidth}{!}{%
\begin{tabular}{@{}lccccccc@{}}
\toprule
Model & PPL $\downarrow$ & MC \% & Gen.\ \% & RC (BPB $\downarrow$) & Basic skills $\downarrow$ & Pattern \% & Pattern BPB $\downarrow$ \\
\midrule
\modelname{} & 30.57 $\pm$ 0.08 & 39.5 $\pm$ 0.1 & 15.0 $\pm$ 0.6 & 1.076 $\pm$ 0.006 & 1.326 $\pm$ 0.018 & 53.2 $\pm$ 0.7 & 1.339 $\pm$ 0.045 \\
\modelname{} w/o states & 32.39 $\pm$ 0.07 & 38.8 $\pm$ 0.2 & 11.9 $\pm$ 0.5 & 1.105 $\pm$ 0.005 & 1.364 $\pm$ 0.007 & 53.2 $\pm$ 0.7 & 1.380 $\pm$ 0.046 \\
Transformer, token-match & 32.44 $\pm$ 0.02 & 38.4 $\pm$ 0.5 & 13.0 $\pm$ 0.6 & 1.112 $\pm$ 0.004 & 1.390 $\pm$ 0.011 & 50.7 $\pm$ 2.0 & 1.427 $\pm$ 0.051 \\
Transformer, compute-match & 31.04 $\pm$ 0.00 & 39.2 $\pm$ 0.3 & 13.6 $\pm$ 0.6 & 1.085 $\pm$ 0.005 & 1.348 $\pm$ 0.036 & 52.2 $\pm$ 0.5 & 1.393 $\pm$ 0.003 \\
Distillation & 31.81 $\pm$ 0.04 & 39.2 $\pm$ 0.4 & 13.0 $\pm$ 0.4 & 1.096 $\pm$ 0.006 & 1.361 $\pm$ 0.019 & 50.5 $\pm$ 1.4 & 1.373 $\pm$ 0.022 \\
\bottomrule
\end{tabular}}
\end{table}

\subsection{Arithmetic by Number of Operands}
\label{app:res-arith}

Tables~\ref{tab:arith-by-length} and~\ref{tab:arith-acc-by-length} break the arithmetic results down by the number of operands: bits per byte of the answer, and exact-match accuracy of the teacher-forced argmax.

\begin{table}[!htbp]
\centering
\scriptsize
\setlength{\tabcolsep}{3pt}
\caption{Arithmetic by number of operands: bits per byte of the answer (lower is better), for the rows of Tables~\ref{tab:lm} and~\ref{tab:teachers} plus the vanilla Transformer trained on 2$\times$ the tokens (Figure~\ref{fig:arith_bpb}); 1{,}000 expressions per operand count (200 at two operands), three-shot. Avg.\ is over all expressions (total bits over total answer bytes), the number reported in Table~\ref{tab:lm}. Bold: best in the column within a size.}
\label{tab:arith-by-length}
\begin{tabular}{@{}cllccccccccccc@{}}
\toprule
\multicolumn{4}{c}{} & \multicolumn{9}{c}{Number of operands $n$} & \\
\cmidrule(lr){5-13}
 & Model & Tokens & \begin{tabular}[c]{@{}c@{}}Teacher\\(size@tokens)\end{tabular} & 2 & 3 & 4 & 5 & 6 & 7 & 8 & 9 & 10 & Avg. \\
\midrule
\multirow{8}{*}{\rotatebox[origin=c]{90}{135M}} & \modelname{} & 13.4B & 135M@29.6B & 2.03 & \textbf{2.13} & \textbf{2.12} & 2.27 & 2.27 & 2.35 & 2.36 & 2.42 & 2.42 & \textbf{2.29} \\
 & \modelname{} & 13.4B & 350M@38.4B & 2.23 & 2.21 & 2.17 & \textbf{2.26} & \textbf{2.25} & \textbf{2.32} & \textbf{2.35} & \textbf{2.38} & \textbf{2.41} & \textbf{2.29} \\
 & \modelname{} w/o states & 13.4B & 135M@29.6B & 2.24 & 2.32 & 2.26 & 2.37 & 2.37 & 2.47 & 2.49 & 2.55 & 2.56 & 2.42 \\
 & \modelname{} w/o states & 13.4B & 350M@38.4B & 2.33 & 2.28 & 2.24 & 2.44 & 2.45 & 2.58 & 2.59 & 2.63 & 2.64 & 2.48 \\
 & Transformer, token-match & 13.4B & -- & \textbf{2.02} & 2.28 & 2.29 & 2.42 & 2.45 & 2.55 & 2.57 & 2.58 & 2.59 & 2.46 \\
 & Transformer, compute-match & 19.3B & -- & 2.03 & 2.21 & 2.32 & 2.48 & 2.52 & 2.61 & 2.62 & 2.62 & 2.65 & 2.50 \\
 & Transformer, 2$\times$ tokens & 29.6B & -- & 2.20 & 2.28 & 2.29 & 2.37 & 2.37 & 2.44 & 2.45 & 2.47 & 2.50 & 2.39 \\
 & Distillation & 13.4B & 135M@29.6B & 2.03 & 2.18 & 2.22 & 2.32 & 2.33 & 2.43 & 2.45 & 2.48 & 2.49 & 2.36 \\
\midrule
\multirow{8}{*}{\rotatebox[origin=c]{90}{350M}} & \modelname{} & 34.9B & 350M@77.8B & 1.56 & 1.97 & 2.00 & \textbf{2.11} & \textbf{2.13} & \textbf{2.18} & \textbf{2.19} & \textbf{2.24} & \textbf{2.24} & \textbf{2.12} \\
 & \modelname{} & 34.9B & 1B@111.8B & \textbf{1.50} & \textbf{1.94} & \textbf{1.99} & \textbf{2.11} & 2.16 & \textbf{2.18} & 2.22 & 2.27 & 2.28 & 2.14 \\
 & \modelname{} w/o states & 34.9B & 350M@77.8B & 1.64 & 2.00 & 2.07 & 2.20 & 2.21 & 2.28 & 2.30 & 2.35 & 2.36 & 2.21 \\
 & \modelname{} w/o states & 34.9B & 1B@111.8B & 1.55 & 2.10 & 2.16 & 2.30 & 2.33 & 2.36 & 2.39 & 2.43 & 2.44 & 2.30 \\
 & Transformer, token-match & 34.9B & -- & 2.04 & 2.27 & 2.24 & 2.38 & 2.45 & 2.54 & 2.59 & 2.60 & 2.62 & 2.46 \\
 & Transformer, compute-match & 49.9B & -- & 1.84 & 2.06 & 2.11 & 2.25 & 2.33 & 2.42 & 2.49 & 2.49 & 2.51 & 2.33 \\
 & Transformer, 2$\times$ tokens & 77.8B & -- & 1.64 & 2.00 & 2.06 & 2.15 & 2.24 & 2.30 & 2.38 & 2.40 & 2.44 & 2.24 \\
 & Distillation & 34.9B & 350M@77.8B & 1.83 & 2.07 & 2.08 & 2.20 & 2.20 & 2.27 & 2.31 & 2.36 & 2.41 & 2.23 \\
\midrule
\multirow{10}{*}{\rotatebox[origin=c]{90}{1B}} & \modelname{} & 107.4B & 1B@222B & 0.74 & 1.86 & 1.93 & 1.96 & 1.99 & 2.04 & 2.07 & 2.11 & 2.13 & 1.99 \\
 & \modelname{} & 107.4B & 1B@4001B & \textbf{0.71} & 1.84 & 1.91 & 1.98 & 2.02 & 2.07 & 2.12 & 2.18 & 2.21 & 2.02 \\
 & \modelname{} & 107.4B & 7B@701B & 0.79 & 1.79 & \textbf{1.83} & \textbf{1.88} & \textbf{1.92} & \textbf{1.98} & \textbf{2.02} & \textbf{2.05} & \textbf{2.06} & \textbf{1.92} \\
 & \modelname{} w/o states & 107.4B & 1B@222B & 0.93 & 1.98 & 1.98 & 2.01 & 2.03 & 2.08 & 2.12 & 2.18 & 2.19 & 2.05 \\
 & \modelname{} w/o states & 107.4B & 1B@4001B & 0.77 & 1.91 & 1.94 & 1.98 & 1.99 & 2.03 & 2.07 & 2.11 & 2.12 & 1.99 \\
 & \modelname{} w/o states & 107.4B & 7B@701B & 1.04 & 1.87 & 1.88 & 1.93 & 1.96 & 2.01 & 2.05 & 2.08 & 2.10 & 1.97 \\
 & Transformer, token-match & 107.4B & -- & 1.04 & 1.86 & 1.97 & 2.01 & 2.03 & 2.07 & 2.12 & 2.15 & 2.16 & 2.03 \\
 & Transformer, compute-match & 152.4B & -- & 0.95 & 1.94 & 2.01 & 2.02 & 2.05 & 2.11 & 2.17 & 2.20 & 2.21 & 2.07 \\
 & Transformer, 2$\times$ tokens & 222B & -- & 0.78 & 1.85 & 1.94 & 1.96 & 2.03 & 2.06 & 2.10 & 2.15 & 2.15 & 2.01 \\
 & Distillation & 107.4B & 1B@222B & 0.77 & \textbf{1.73} & 1.92 & 1.93 & 1.95 & 2.00 & 2.05 & 2.07 & 2.09 & 1.95 \\
\bottomrule
\end{tabular}
\end{table}

\begin{table}[!htbp]
\centering
\scriptsize
\setlength{\tabcolsep}{3pt}
\caption{Arithmetic by number of operands: exact-match accuracy \% of the teacher-forced argmax, for the expressions and rows of Table~\ref{tab:arith-by-length}. Avg.\ is over all expressions. Bold: best in the column within a size.}
\label{tab:arith-acc-by-length}
\begin{tabular}{@{}cllccccccccccc@{}}
\toprule
\multicolumn{4}{c}{} & \multicolumn{9}{c}{Number of operands $n$} & \\
\cmidrule(lr){5-13}
 & Model & Tokens & \begin{tabular}[c]{@{}c@{}}Teacher\\(size@tokens)\end{tabular} & 2 & 3 & 4 & 5 & 6 & 7 & 8 & 9 & 10 & Avg. \\
\midrule
\multirow{8}{*}{\rotatebox[origin=c]{90}{135M}} & \modelname{} & 13.4B & 135M@29.6B & \textbf{7.0} & 5.0 & 3.4 & 2.4 & 3.2 & 2.5 & 2.2 & 3.0 & 1.9 & 3.0 \\
 & \modelname{} & 13.4B & 350M@38.4B & 5.5 & 4.4 & \textbf{4.4} & \textbf{3.3} & \textbf{4.0} & 2.5 & 2.2 & 2.5 & \textbf{3.6} & \textbf{3.4} \\
 & \modelname{} w/o states & 13.4B & 135M@29.6B & 6.0 & 4.5 & 3.8 & 3.1 & 3.1 & 2.7 & 1.7 & \textbf{3.1} & 2.8 & 3.2 \\
 & \modelname{} w/o states & 13.4B & 350M@38.4B & 5.0 & 4.7 & 3.5 & \textbf{3.3} & 3.9 & 2.0 & 2.3 & 2.8 & 2.3 & 3.1 \\
 & Transformer, token-match & 13.4B & -- & \textbf{7.0} & 4.5 & 4.1 & 3.0 & 2.6 & \textbf{3.0} & 2.4 & 2.0 & 2.9 & 3.2 \\
 & Transformer, compute-match & 19.3B & -- & 5.5 & 4.6 & 3.3 & 2.2 & 2.9 & 2.7 & \textbf{3.0} & 3.0 & 2.0 & 3.0 \\
 & Transformer, 2$\times$ tokens & 29.6B & -- & 4.5 & 4.6 & 3.4 & 2.3 & 2.8 & \textbf{3.0} & 2.8 & 2.6 & 2.8 & 3.1 \\
 & Distillation & 13.4B & 135M@29.6B & \textbf{7.0} & \textbf{5.2} & 3.9 & 2.9 & 2.6 & 2.2 & 2.5 & 2.1 & 2.4 & 3.1 \\
\midrule
\multirow{8}{*}{\rotatebox[origin=c]{90}{350M}} & \modelname{} & 34.9B & 350M@77.8B & 21.5 & \textbf{6.4} & 4.2 & \textbf{4.2} & 2.4 & 2.7 & \textbf{3.3} & 2.5 & 2.5 & \textbf{4.0} \\
 & \modelname{} & 34.9B & 1B@111.8B & 22.5 & 5.6 & \textbf{5.3} & 4.1 & 2.5 & 2.2 & 3.0 & 3.1 & 1.9 & 3.9 \\
 & \modelname{} w/o states & 34.9B & 350M@77.8B & 15.0 & \textbf{6.4} & 4.3 & 3.6 & 3.1 & 1.9 & 2.2 & 3.1 & 2.6 & 3.7 \\
 & \modelname{} w/o states & 34.9B & 1B@111.8B & \textbf{23.0} & 5.4 & 4.9 & 3.8 & 2.0 & 2.5 & 2.5 & 2.9 & 2.3 & 3.8 \\
 & Transformer, token-match & 34.9B & -- & 9.5 & 5.1 & 4.3 & 3.4 & 2.8 & 3.0 & 2.2 & 2.2 & \textbf{3.1} & 3.4 \\
 & Transformer, compute-match & 49.9B & -- & 8.5 & 4.9 & 4.6 & 2.9 & 1.9 & \textbf{3.3} & 2.7 & 2.5 & 2.2 & 3.3 \\
 & Transformer, 2$\times$ tokens & 77.8B & -- & 13.0 & 6.1 & 3.3 & 3.2 & 2.9 & 2.7 & 2.9 & \textbf{3.4} & 2.7 & 3.6 \\
 & Distillation & 34.9B & 350M@77.8B & 11.0 & 5.1 & 4.1 & 3.8 & \textbf{3.4} & 2.1 & 2.7 & 2.9 & 2.8 & 3.5 \\
\midrule
\multirow{10}{*}{\rotatebox[origin=c]{90}{1B}} & \modelname{} & 107.4B & 1B@222B & 58.5 & 8.7 & 5.3 & 5.2 & \textbf{4.9} & 3.7 & 3.0 & 3.0 & 2.6 & 5.9 \\
 & \modelname{} & 107.4B & 1B@4001B & 63.0 & 12.8 & 4.4 & 4.5 & 4.3 & \textbf{4.5} & 3.4 & 2.6 & 2.0 & 6.2 \\
 & \modelname{} & 107.4B & 7B@701B & \textbf{63.5} & 11.4 & 5.7 & 4.5 & 3.7 & 2.9 & \textbf{3.7} & 3.6 & 2.5 & 6.2 \\
 & \modelname{} w/o states & 107.4B & 1B@222B & 46.5 & 8.9 & 4.6 & 4.5 & 4.1 & 2.8 & 2.6 & 1.9 & 2.3 & 5.0 \\
 & \modelname{} w/o states & 107.4B & 1B@4001B & 59.5 & 10.8 & 4.4 & 3.9 & 4.1 & 3.6 & 3.0 & 3.1 & \textbf{3.5} & 5.9 \\
 & \modelname{} w/o states & 107.4B & 7B@701B & 47.0 & 10.8 & 5.3 & 4.3 & 2.9 & 2.7 & 3.4 & \textbf{4.5} & 3.0 & 5.6 \\
 & Transformer, token-match & 107.4B & -- & 46.5 & 8.1 & 4.6 & 4.1 & 4.1 & 2.1 & 3.3 & 4.0 & 2.2 & 5.1 \\
 & Transformer, compute-match & 152.4B & -- & 48.5 & 9.0 & 3.8 & 3.5 & 3.7 & 2.0 & 3.0 & 3.4 & 2.2 & 4.9 \\
 & Transformer, 2$\times$ tokens & 222B & -- & 56.0 & 13.3 & 4.1 & \textbf{5.4} & 3.8 & 3.0 & 2.4 & 3.3 & 2.3 & 6.0 \\
 & Distillation & 107.4B & 1B@222B & 56.5 & \textbf{14.1} & \textbf{6.2} & 5.1 & 4.8 & 2.5 & 2.4 & 3.4 & 2.9 & \textbf{6.4} \\
\bottomrule
\end{tabular}
\end{table}

\subsection{Pattern Continuation}
\label{app:res-pattern}

Table~\ref{tab:pattern} gives the pattern-continuation results discussed in \S\ref{sec: results}.

\begin{table}[!htbp]
\centering
\small
\setlength{\tabcolsep}{5pt}
\caption{Pattern continuation, the basic-skills task of OLMES (534 five-shot questions, e.g., \texttt{2 4 6 8} \(\to\) \texttt{10}): cloze accuracy \% and bits per byte of the answer (BPB, lower is better), for the rows of Table~\ref{tab:lm} at the paper's training seed, and the Transformer trained on 2$\times$ the tokens (gray, outside the bold rule). Three-seed statistics at 135M are in Table~\ref{tab:olmes-seeds}. Bold: best in column.}
\label{tab:pattern}
\begin{tabular}{@{}lcccccc@{}}
\toprule
 & \multicolumn{2}{c}{135M} & \multicolumn{2}{c}{350M} & \multicolumn{2}{c}{1B} \\
\cmidrule(lr){2-3}\cmidrule(lr){4-5}\cmidrule(l){6-7}
Model & Acc.\ \% & BPB $\downarrow$ & Acc.\ \% & BPB $\downarrow$ & Acc.\ \% & BPB $\downarrow$ \\
\midrule
\modelname{} & \textbf{53.2} & \textbf{1.356} & \textbf{64.4} & 1.134 & \textbf{72.1} & \textbf{0.851} \\
\modelname{} w/o states & 52.6 & 1.383 & 63.7 & \textbf{1.121} & 70.6 & 0.965 \\
Transformer, token-match & 50.7 & 1.443 & 58.4 & 1.216 & 66.1 & 0.948 \\
Transformer, compute-match & 52.6 & 1.396 & 58.1 & 1.144 & 67.6 & 0.955 \\
Distillation & 48.9 & 1.392 & 60.7 & 1.153 & 69.7 & 0.890 \\
\textcolor{black!65}{Transformer, 2$\times$ tokens} & \textcolor{black!65}{53.6} & \textcolor{black!65}{1.352} & \textcolor{black!65}{63.5} & \textcolor{black!65}{1.180} & \textcolor{black!65}{70.4} & \textcolor{black!65}{0.896} \\
\bottomrule
\end{tabular}
\end{table}

\subsection{Feedback-Transformer Comparison: Results}
\label{app:jacobi-results}

\paragraph{Seeds.}
Every model of Table~\ref{tab:jacobi} except T\textsuperscript{2}MLR (HF) is trained with three seeds (42, 43, 44). Table~\ref{tab:jacobi-seeds} reports the mean and standard deviation of each column; Table~\ref{tab:jacobi} shows the means.

\begin{table}[!htbp]
\centering
\small
\caption{The models of Table~\ref{tab:jacobi} over three training seeds (42 / 43 / 44): mean $\pm$ sd of each column. T\textsuperscript{2}MLR (HF) is not listed: it is the authors' single released checkpoint, so it has no seed statistics. We regard a difference between two models as beyond seed noise when their mean $\pm$ sd ranges do not overlap. T\textsuperscript{2}MLR (our run) sets the compute budget, so its row is the same under both budgets. Bold: the best mean within each budget. Per-suite results of seed 42 are in Table~\ref{tab:olmes-jacobi}.}
\label{tab:jacobi-seeds}
\resizebox{\linewidth}{!}{%
\begin{tabular}{@{}lccccc@{}}
\toprule
Model & PPL $\downarrow$ & Arith.\ $\downarrow$ & MC \% & Gen.\ \% & RC $\downarrow$ \\
\midrule
\multicolumn{6}{@{}l}{\emph{Token-matched (10B tokens)}} \\
\modelname{} & \textbf{19.42} $\pm$ 0.02 & \textbf{2.182} $\pm$ 0.007 & \textbf{36.7} $\pm$ 0.27 & \textbf{7.8} $\pm$ 0.61 & \textbf{1.705} $\pm$ 0.023 \\
Transformer & 20.30 $\pm$ 0.03 & 2.345 $\pm$ 0.156 & 36.2 $\pm$ 0.19 & 6.7 $\pm$ 0.21 & 1.761 $\pm$ 0.002 \\
T\textsuperscript{2}MLR (our run) & 19.92 $\pm$ 0.06 & 2.267 $\pm$ 0.101 & 36.5 $\pm$ 0.04 & \textbf{7.8} $\pm$ 0.94 & 1.737 $\pm$ 0.010 \\
Multi-pass Transformer & 20.40 $\pm$ 0.04 & 2.249 $\pm$ 0.064 & 36.3 $\pm$ 0.36 & 6.5 $\pm$ 0.25 & 1.773 $\pm$ 0.017 \\
\midrule
\multicolumn{6}{@{}l}{\emph{Compute-matched ($2.3{\times}10^{19}$ FLOPs)}} \\
\modelname{} & 18.04 $\pm$ 0.04 & \textbf{2.184} $\pm$ 0.032 & 38.0 $\pm$ 0.19 & \textbf{9.5} $\pm$ 0.33 & 1.629 $\pm$ 0.016 \\
Transformer & \textbf{18.02} $\pm$ 0.02 & 2.296 $\pm$ 0.058 & \textbf{38.1} $\pm$ 0.37 & 8.5 $\pm$ 0.23 & \textbf{1.622} $\pm$ 0.008 \\
T\textsuperscript{2}MLR (our run) & 19.92 $\pm$ 0.06 & 2.267 $\pm$ 0.101 & 36.5 $\pm$ 0.04 & 7.8 $\pm$ 0.94 & 1.737 $\pm$ 0.010 \\
Multi-pass Transformer & 18.58 $\pm$ 0.03 & 2.249 $\pm$ 0.037 & 37.6 $\pm$ 0.02 & 7.7 $\pm$ 0.17 & 1.675 $\pm$ 0.008 \\
\bottomrule
\end{tabular}}
\end{table}

Table~\ref{tab:olmes-jacobi} gives every OLMES suite for \S\ref{sec:jacobi} (seed 42; three-seed statistics in Table~\ref{tab:jacobi-seeds}).

\begin{table}[!htbp]
\centering
\scriptsize
\setlength{\tabcolsep}{2.5pt}
\caption{OLMES results for the feedback-Transformer comparison (Table~\ref{tab:jacobi}), every suite of the scan; the T\textsuperscript{2}MLR columns coincide across the two budgets. Rows and units as in Table~\ref{tab:olmes-135m}. Bold: best within a budget.}
\label{tab:olmes-jacobi}
\begin{tabular}{@{}lcccccccccc@{}}
\toprule
& \multicolumn{5}{c}{\textbf{Token-matched} \textcolor{gray}{10B tokens}} & \multicolumn{5}{c}{\textbf{Compute-matched} \textcolor{gray}{$2.3{\times}10^{19}$ FLOPs}} \\
\cmidrule(lr){2-6}\cmidrule(l){7-11}
Suite & \begin{tabular}[b]{@{}c@{}}\modelname{}\end{tabular} & \begin{tabular}[b]{@{}c@{}}Trans-\\former\end{tabular} & \begin{tabular}[b]{@{}c@{}}T\textsuperscript{2}MLR\\(ours)\end{tabular} & \begin{tabular}[b]{@{}c@{}}T\textsuperscript{2}MLR\\(HF)\end{tabular} & \begin{tabular}[b]{@{}c@{}}Multi-\\pass\end{tabular} & \begin{tabular}[b]{@{}c@{}}\modelname{}\end{tabular} & \begin{tabular}[b]{@{}c@{}}Trans-\\former\end{tabular} & \begin{tabular}[b]{@{}c@{}}T\textsuperscript{2}MLR\\(ours)\end{tabular} & \begin{tabular}[b]{@{}c@{}}T\textsuperscript{2}MLR\\(HF)\end{tabular} & \begin{tabular}[b]{@{}c@{}}Multi-\\pass\end{tabular} \\
\midrule
Core 8 (cloze) & 42.0 & 40.7 & \textbf{42.5} & 40.7 & 41.9 & 43.8 & \textbf{44.1} & 42.5 & 40.7 & 43.4 \\
MMLU (cloze) & 27.4 & \textbf{27.5} & 27.2 & 26.5 & 26.9 & 27.9 & \textbf{28.3} & 27.2 & 26.5 & 27.9 \\
Basic skills (cloze) & \textbf{40.4} & 39.8 & 39.6 & 38.7 & 40.1 & 42.1 & \textbf{42.9} & 39.6 & 38.7 & 41.4 \\
Generative QA & 5.2 & 4.5 & \textbf{5.7} & 3.8 & 4.6 & \textbf{8.0} & 7.8 & 5.7 & 3.8 & 6.6 \\
TriviaQA & 4.1 & \textbf{4.2} & 4.0 & \textbf{4.2} & 3.6 & 3.9 & 3.7 & 4.0 & \textbf{4.2} & \textbf{4.2} \\
BBH (CoT) & \textbf{13.9} & 10.8 & 13.7 & 11.6 & 11.7 & \textbf{17.5} & 14.7 & 13.7 & 11.6 & 12.7 \\
ARC BPB & \textbf{1.047} & 1.078 & 1.077 & 1.110 & 1.067 & 1.004 & \textbf{0.993} & 1.077 & 1.110 & 1.018 \\
MMLU BPB & \textbf{1.359} & 1.390 & 1.384 & 1.442 & 1.392 & \textbf{1.319} & 1.321 & 1.384 & 1.442 & 1.335 \\
Basic skills BPB & \textbf{1.593} & 1.680 & 1.637 & 1.744 & 1.671 & 1.521 & \textbf{1.515} & 1.637 & 1.744 & 1.546 \\
MATH BPB & \textbf{1.864} & 1.944 & 1.937 & 2.011 & 2.015 & \textbf{1.808} & 1.877 & 1.937 & 2.011 & 1.943 \\
HumanEval BPB & \textbf{2.121} & 2.268 & 2.161 & 2.273 & 2.188 & 2.050 & \textbf{2.044} & 2.161 & 2.273 & 2.093 \\
MBPP BPB & \textbf{2.091} & 2.215 & 2.176 & 2.319 & 2.201 & \textbf{1.964} & 1.967 & 2.176 & 2.319 & 2.077 \\
\midrule
\multicolumn{11}{l}{\emph{Not in Table~\ref{tab:jacobi}}} \\
BoolQ (cloze) & 61.1 & 56.2 & 44.1 & \textbf{63.9} & 46.9 & 63.8 & 61.6 & 44.1 & \textbf{63.9} & 58.7 \\
MMLU (letter) & 25.3 & 25.2 & 26.0 & 25.2 & \textbf{27.0} & 25.4 & 25.8 & \textbf{26.0} & 25.2 & 25.8 \\
AGI-Eval (letter) & 21.4 & \textbf{23.6} & 22.3 & 22.9 & 22.7 & 20.7 & 21.4 & 22.3 & 22.9 & \textbf{23.5} \\
GSM8K (5-shot) & 2.4 & \textbf{2.6} & 1.7 & 1.4 & 1.7 & \textbf{2.6} & 2.4 & 1.7 & 1.4 & 2.1 \\
GSM8K (8-shot) & \textbf{2.4} & 2.0 & 1.4 & 1.6 & 1.7 & \textbf{2.4} & 1.7 & 1.4 & 1.6 & 1.8 \\
MATH-500 & \textbf{0.0} & \textbf{0.0} & \textbf{0.0} & \textbf{0.0} & \textbf{0.0} & \textbf{0.0} & \textbf{0.0} & \textbf{0.0} & \textbf{0.0} & \textbf{0.0} \\
Minerva MATH & 0.0 & 0.0 & \textbf{0.1} & 0.0 & 0.0 & 0.0 & \textbf{0.1} & \textbf{0.1} & 0.0 & 0.0 \\
HumanEval pass@1 & \textbf{0.0} & \textbf{0.0} & \textbf{0.0} & \textbf{0.0} & \textbf{0.0} & \textbf{0.0} & \textbf{0.0} & \textbf{0.0} & \textbf{0.0} & \textbf{0.0} \\
MBPP pass@1 & \textbf{0.0} & \textbf{0.0} & \textbf{0.0} & \textbf{0.0} & \textbf{0.0} & \textbf{0.0} & \textbf{0.0} & \textbf{0.0} & \textbf{0.0} & \textbf{0.0} \\
\bottomrule
\end{tabular}
\end{table}

\FloatBarrier

\section{Ablations}
\label{app:ablations}

Beyond a Transformer, \modelname{} spends about one forward pass per training token, the teacher's. The compute-matched Transformer spends the same compute on over \(40\%\) more tokens instead, and \modelname{} is ahead of it at every scale (Table~\ref{tab:lm}). Here we ask where that lead comes from, changing one thing at a time: how the teacher's forward pass is used (\S\ref{app:teacher-ablation}), how wide the fed-back state is (\S\ref{app:k-sweep}), whether the gain is built in the teacher-forced trunk or in the adaptation phase (\S\ref{app:trunk-adaptation}), and how strong the teacher must be (\S\ref{app:teacher-strength}). Every variant keeps the architecture, recipe, learning-rate schedule and data order of \S\ref{app:impl}. Figures~\ref{fig:teacher-ablation} and~\ref{fig:k-sweep} plot each variant's improvement over the compute-matched Transformer, so the zero line stands for spending the extra pass on more data. We lead with perplexity on the full C4 validation subset, which separates the variants well beyond seed noise, and report the downstream columns of Table~\ref{tab:lm} after it.

\subsection{Teacher Forcing: Inputs, Targets, or No Teacher}
\label{app:teacher-ablation}

We compare four ways to spend the extra forward pass:
\begin{enumerate}[leftmargin=*,itemsep=2pt,topsep=3pt,parsep=0pt]
    \item \textbf{\modelname{}}: the teacher's states are both the inputs and the alignment target (\S\ref{sec:training}).
    \item \textbf{\modelname{}, teacher as input only}: \modelname{} with \(\lambda=0\) throughout, so the teacher's states are inputs and never targets.
    \item \textbf{\modelname{}, no teacher (1 Jacobi iteration)}: no teacher at any point; the extra pass is the model's own. Each training step first runs a pass without gradient in which every position after the first receives the bias \(\mathbf b\), and the states it predicts (Eq.~\ref{eq:state}) are fed to the pass that carries the loss, with prefix state dropout as usual and \(\lambda=0\). This is the adaptation phase (\S\ref{app:adaptation}) run from the first step, one Jacobi iteration without gradient between the passes. The last \(10\%\) of training is \modelname{}'s adaptation phase, unchanged.
    \item \textbf{Distillation}: the teacher's states are the target only, and nothing is fed back (the baseline of Table~\ref{tab:lm}).
\end{enumerate}
Every variant is trained at 135M with the three seeds of Table~\ref{tab:olmes-seeds}.

\begin{figure}[ht]
    \centering
    \includegraphics[width=\linewidth]{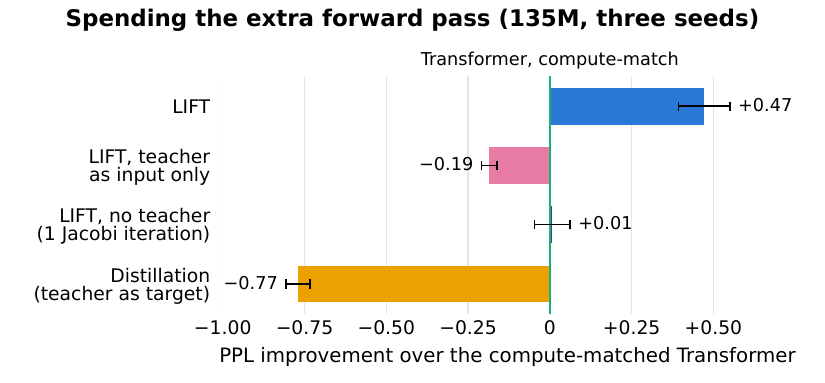}
    \caption{Teacher ablation at 135M: spending \modelname{}'s extra forward pass. Each bar is a model's perplexity improvement on the full C4 validation subset over the compute-matched Transformer, which spends that compute on more tokens (the zero line; right is better): the mean over three training seeds of the per-seed difference, with whiskers of \(\pm1\) sd.}
    \label{fig:teacher-ablation}
\end{figure}

\paragraph{Only the teacher's states as inputs and targets beat the compute-matched Transformer.}
At 135M, \modelname{} improves perplexity over the compute-matched Transformer by \(0.47\pm0.08\) (Figure~\ref{fig:teacher-ablation}). Trained on its own states, the same architecture ties it (\(+0.01\pm0.05\), within \(0.07\) at every seed); the teacher thus accounts for \(98\%\) of \modelname{}'s gain over the compute-matched Transformer. Self-generated states do improve on a Transformer trained on the same tokens, by \(1.40\) perplexity, but spending the extra pass on data instead improves on it by about as much, \(1.39\). With the teacher's states as inputs only (\(\lambda=0\)), \modelname{} falls \(0.19\pm0.02\) behind the compute-matched Transformer (\(31.23\pm0.03\) against \(31.04\pm0.00\)), and behind the no-teacher variant too. Without the alignment term the model makes less use of the teacher's states, reaching \(30.72\pm0.04\) when fed them against \(30.25\pm0.05\) for \modelname{}, and its own states drift further from them: feeding it its own states instead costs \(0.51\) perplexity, against \(0.32\) for \modelname{}, and run without states it is a full point behind the token-matched Transformer (\(33.47\) against \(32.44\)). The alignment term is thus what carries the teacher-forced channel over to inference (\S\ref{sec:training}).

\paragraph{As a target only, the teacher is worth less than the data.}
Distillation falls behind the compute-matched Transformer by \(0.77\pm0.04\). Its forward pass through the teacher is better spent on more tokens; used as both inputs and targets, the same teacher's states give \modelname{} its lead.

\paragraph{Tuned distillation.}
The Distillation baseline of Table~\ref{tab:lm} reuses \modelname{}'s support and temperature: the top $1{,}024$ tokens with an aggregated tail, $\tau=1.5$, $\lambda=1.5$, next-token loss kept. Three further variants at 135M, with the same teacher, seed, data order and schedule, do not close the gap. The same baseline at $\tau=1$ ties it (31.80 against 31.77). The recipe of \citet{busbridge2025distillation}, a pure forward KL over the full vocabulary at $\tau=1$ without the next-token loss, is behind even the token-matched Transformer (32.66 against 32.43, $[0.19,0.29]$), and annealing that student with the next-token loss recovers only part of the gap (32.29). \modelname{} leads the best of the four by $1.11$ ($[1.06,1.17]$).

\paragraph{Downstream columns.}
At 135M and matched compute, most downstream differences are within seed noise, in the sense of Table~\ref{tab:olmes-seeds} (overlapping mean \(\pm\) sd ranges). Two are beyond it: \modelname{}'s generative score (\(15.0\pm0.6\) against \(13.6\pm0.6\)) and Distillation's reference-completion loss (\(1.096\pm0.006\) against \(1.085\pm0.005\) bits per byte). Without a teacher no column moves beyond noise; on arithmetic the no-teacher variant has \modelname{}'s mean (\(2.28\) bits per byte for both), and both are within the compute-matched Transformer's spread (\(2.39\pm0.09\)).

\paragraph{Backpropagating through own states.}
Training on the model's own states exactly means backpropagating through the recurrence, as \modelname{}-RNN does on \(S_5\) (\S\ref{sec:synthetic}). This makes training sequential over positions, removing the parallelism that pretraining relies on: each position waits for the state of the one before it, the backward pass traverses every step, and the attention inputs of every step must be stored, which grows quadratically with the sequence length in a direct implementation; activation checkpointing and fused attention kernels, built for parallel training, do not apply as they are. Feedback models therefore truncate the recurrence to a few parallel passes and backpropagate through them, as T\textsuperscript{2}MLR and the Multi-pass Transformer do. At matched compute both fall behind the vanilla Transformer, and both need more memory than \modelname{} (\S\ref{sec:jacobi}, Table~\ref{tab:jacobi-footprint}). The no-teacher variant above is the cheapest point of this family, one extra pass without gradient, and ties the compute-matched Transformer. None of the variants that generate their own training states beats a Transformer given the same compute; \modelname{}, whose training states come from the teacher, does.

\subsection{Channel Width}
\label{app:k-sweep}

\begin{figure}[ht]
    \centering
    \includegraphics[width=\linewidth]{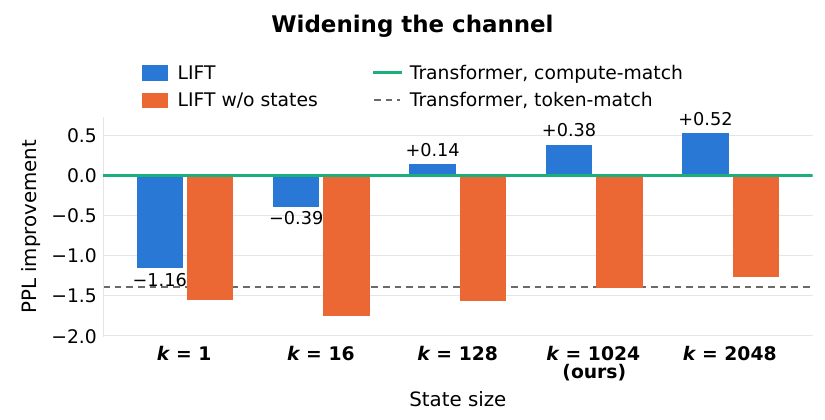}
    \caption{Channel width at 135M: improvement in perplexity over the compute-matched Transformer (the zero line; up is better) for \(k\) from \(1\) to \(2048\), with the model's own states fed back (\modelname{}) and for the same model run without states (\modelname{} w/o states). Dashed: the token-matched Transformer. Every step in \(k\) is beyond the paired bootstrap \(95\%\) interval over evaluation sequences (Table~\ref{tab:ppl-bootstrap}'s protocol).}
    \label{fig:k-sweep}
\end{figure}

We vary \(k\), the number of teacher tokens a state keeps (Eq.~\ref{eq:state}), at 135M, in both the fed-back state and the support of the alignment term, with everything else as in Table~\ref{tab:lm}, including its seed. At \(k=1\) the state is the teacher's top token scaled by its probability, a second token channel; \(k=1024\) is our setting. Figure~\ref{fig:k-sweep} shows each model with its states and, for reference, the same model run without them (\modelname{} w/o states in Table~\ref{tab:lm}).

\paragraph{The gain grows with the width of the channel.}
Perplexity improves monotonically with \(k\). The one-token channel is \(1.16\) behind the compute-matched Transformer and \(k=16\) is \(0.39\) behind; only \(k\geq128\) is ahead of it, by \(0.14\), by \(0.38\) at \(k=1024\) and by \(0.52\) at \(k=2048\). Measured against the token-matched Transformer, \(k=1\) keeps \(13\%\) of \modelname{}'s gain, \(k=16\) keeps \(56\%\) and \(k=128\) keeps \(86\%\). An extra token of input is thus not what \modelname{}'s states contribute; their width is. Arithmetic follows the same order: \(2.50\), \(2.35\), \(2.31\) and \(2.29\) bits per byte at \(k=1\), \(16\), \(128\) and \(1024\), the one-token channel at the level of the Transformers (\(2.46\) token-matched, \(2.50\) compute-matched).

\paragraph{The width acts through the states.}
Run without states, the models trained at every \(k\) are within \(0.48\) perplexity of each other and within \(0.36\) of the token-matched Transformer: widening the channel, which also widens the alignment term's support, does not train a better Transformer. What grows with \(k\) is what the states carry at inference, \(0.40\) perplexity at \(k=1\), \(1.36\) at \(k=16\), \(1.70\) at \(k=128\), \(1.78\) at \(k=1024\) and \(1.79\) at \(k=2048\).

\paragraph{Choice of \(k\).}
Returns diminish with width, from \(0.77\) perplexity between \(k=1\) and \(16\) and \(0.53\) between \(16\) and \(128\) to \(0.24\) between \(128\) and \(1024\), still well beyond evaluation noise (paired bootstrap \(95\%\) interval \([0.20, 0.29]\)). Doubling to \(k=2048\) improves perplexity by a further \(0.14\) (\([0.10, 0.18]\); one training seed, against a seed spread of \(0.08\) for \modelname{} at 135M), but not through the states: what they carry has stopped growing (\(1.78\) at \(k=1024\), \(1.79\) at \(k=2048\)), and the model run without states improves by about as much (\(32.31\) against \(32.44\)), within the spread of the without-states models across \(k\). The width costs almost nothing: \(k\) enters the compute only through the soft-token sum \(E^{\top}\mathbf s_i\), \(0.13\%\) of a training step at 1B for \(k=1024\) (Table~\ref{tab:overhead}). We use \(k=1024\) at every scale.

\subsection{Trunk vs.\ Adaptation Phase}
\label{app:trunk-adaptation}

\paragraph{The gain is built by the teacher-forced trunk, not by the adaptation phase.}
Two further runs isolate the adaptation phase of \S\ref{sec:training}. \emph{Adaptation only}: we take the token-matched Transformer at \(90\%\) of training, add a freshly initialized fusion layer, and run \modelname{}'s adaptation phase over the last \(10\%\). At 135M this reaches \(32.53\) perplexity, slightly worse than the Transformer itself (\(32.43\)); at 1B, \(19.68\) against \(19.79\), \(12\%\) of \modelname{}'s gain over the Transformer (\(18.81\)). \emph{No adaptation}: \modelname{} at 1B trained with the teacher until the end keeps \(73\%\) of that gain (\(19.07\)) and is still ahead of the compute-matched Transformer (\(19.17\)); adapting over the last \(5\%\) instead of \(10\%\) recovers the rest (\(18.81\)). The adaptation phase thus fits to the model's own states a channel that the teacher-forced trunk has already made useful. Run over the whole of training, it is the no-teacher variant above, which only ties the compute-matched Transformer.

\subsection{Teacher Strength}
\label{app:teacher-strength}
Table~\ref{tab:teachers} reports \modelname{} trained at each scale with the same-size teacher of \S\ref{sec:setting} and with stronger teachers, either larger or trained for much longer.

\begin{table}[!ht]
\centering
\small
\setlength{\tabcolsep}{4.5pt}
\renewcommand{\arraystretch}{1.12}
\caption{Teacher choice: \modelname{} and \modelname{} w/o states per teacher (size@training tokens) against the token-matched vanilla Transformer. \emph{Tokens}: the training budget of every model in the row block, $5\times$ the Chinchilla budget of its size. \emph{Teacher PPL}: the teacher's own perplexity on the same evaluation set. The results hold on every metric, and barely change, with a larger teacher or one trained much longer. Other columns as in Table~\ref{tab:lm}. Bold: best in column within a size.}
\label{tab:teachers}
\begin{tabular}{@{}lllccccc@{}}
\toprule
Model & Size & Tokens & \begin{tabular}[c]{@{}c@{}}Teacher\\(size@tokens)\end{tabular} & \begin{tabular}[c]{@{}c@{}}Teacher\\PPL\end{tabular} & PPL $\downarrow$ & Arith.\ $\downarrow$ & MC \% \\
\midrule
\modelname{} & 135M & 13.4B & 135M@29.6B & 29.80 & 30.65 & \textbf{2.29} & 39.4 \\
\modelname{} & 135M & 13.4B & 350M@38.4B & 24.91 & \textbf{30.40} & \textbf{2.29} & \textbf{39.5} \\
\modelname{} w/o states & 135M & 13.4B & 135M@29.6B & 29.80 & 32.44 & 2.42 & 38.7 \\
\modelname{} w/o states & 135M & 13.4B & 350M@38.4B & 24.91 & 32.40 & 2.48 & 39.0 \\
Transformer, token-match & 135M & 13.4B & -- & -- & 32.43 & 2.46 & 38.1 \\
\midrule
\modelname{} & 350M & 34.9B & 350M@77.8B & 23.49 & 23.82 & \textbf{2.12} & 45.0 \\
\modelname{} & 350M & 34.9B & 1B@111.8B & 19.72 & \textbf{23.78} & 2.14 & \textbf{45.3} \\
\modelname{} w/o states & 350M & 34.9B & 350M@77.8B & 23.49 & 24.96 & 2.21 & 44.2 \\
\modelname{} w/o states & 350M & 34.9B & 1B@111.8B & 19.72 & 25.01 & 2.30 & 44.2 \\
Transformer, token-match & 350M & 34.9B & -- & -- & 25.15 & 2.46 & 43.2 \\
\midrule
\modelname{} & 1B & 107.4B & 1B@222B & 18.62 & 18.81 & 1.99 & \textbf{52.5} \\
\modelname{} & 1B & 107.4B & 1B@4001B & 17.31 & \textbf{18.77} & 2.02 & 51.9 \\
\modelname{} & 1B & 107.4B & 7B@701B & 14.97 & 18.85 & \textbf{1.92} & 51.7 \\
\modelname{} w/o states & 1B & 107.4B & 1B@222B & 18.62 & 19.62 & 2.05 & 51.2 \\
\modelname{} w/o states & 1B & 107.4B & 1B@4001B & 17.31 & 19.64 & 1.99 & 50.9 \\
\modelname{} w/o states & 1B & 107.4B & 7B@701B & 14.97 & 19.78 & 1.97 & 50.5 \\
Transformer, token-match & 1B & 107.4B & -- & -- & 19.79 & 2.03 & 50.5 \\
\bottomrule
\end{tabular}
\end{table}

\FloatBarrier

\section{Prefill Modes: Refining the Prompt's Key-Value Cache}
\label{app:prefill}

During prefill nothing is generated: the pass over the prompt only builds the key-value cache that decoding attends to, and the state entering the first answer token. The evaluations of \S\ref{sec:experiments} build this cache sequentially, feeding each prompt position the state predicted at the preceding one. Here we ask whether the cache can be built in parallel instead, and at what price. Decoding is sequential in every mode, as it is in any autoregressive Transformer: each answer token is fed the state predicted at the preceding position (Eq.~\ref{eq:state}). Only the construction of the prompt's cache differs.

\paragraph{Prefill modes.}
\textbf{W/o states}: one parallel pass over the prompt with the learned bias \(\mathbf b\) at every position after the first, as in a standard Transformer. \textbf{{\boldmath\(R\)} refinements}: \(R\) further parallel passes over the prompt, each fed, at every position after the first, the state that the previous pass predicted at the preceding position; the first decoding step attends to the cache of the last pass. \textbf{Sequential}: the prompt is processed one position at a time, as the answer is; this is \modelname{} as evaluated in \S\ref{sec:experiments}. The reference point is \modelname{} w/o states, which uses no states in prefill or decoding.

\paragraph{Relative gain.}
For each group \(g\) of MC, Gen.\ and RC, the \emph{states gain} is \(G_g = x_g^{\modelname{}} - x_g^{\text{w/o}}\), the difference between \modelname{}, with states through prefill and decoding, and \modelname{} w/o states. The relative gain of a mode or model \(X\) is \((x_g^{X}-x_g^{\text{w/o}})/G_g\), and we report its mean over the three groups: \(0\%\) is \modelname{} w/o states and \(100\%\) is \modelname{}.

\begin{figure}[htbp]
\centering
\includegraphics[width=\linewidth]{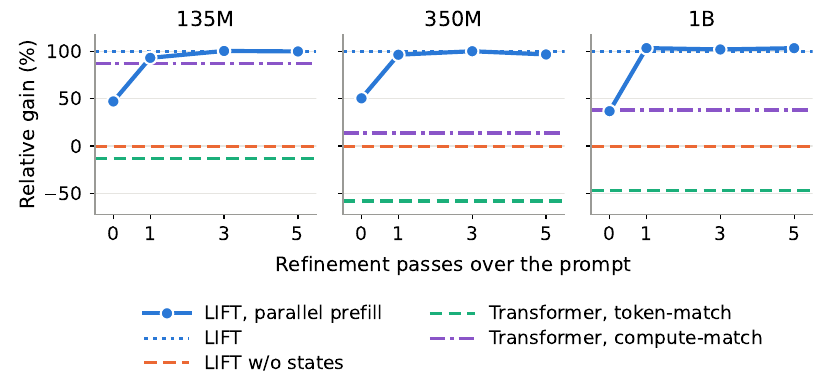}
\caption{Relative gain of \modelname{} with parallel prefill (blue points) by the number of refinement passes over the prompt, against \modelname{} (\(100\%\)) and \modelname{} w/o states (\(0\%\)). The token- and compute-matched Transformers are placed on the same scale; below \(0\%\), a model is behind \modelname{} w/o states.}
\label{fig:prefill}
\end{figure}

\paragraph{One refinement pass suffices for the key-value cache.}
With one refinement pass, \modelname{} matches sequential prefill to within \(0.2\) points on MC, Gen.\ and RC at every scale, a relative gain of \(93\)--\(104\%\); three and five passes change the scores by at most \(0.2\) points (Figure~\ref{fig:prefill}, Table~\ref{tab:prefill}). Arithmetic, reported separately in Table~\ref{tab:prefill}, follows the same pattern: one refinement recovers \(89\)--\(102\%\) of its states gain, and three passes essentially all of it. The Full-bandwidth Transformer shows the same pattern for its fused prefill passes, where most of the improvement comes from the first one~\citep{wang2026fullbandwidth}. Refining the prompt's cache thus costs one extra parallel forward pass rather than \(n\) sequential steps.

\paragraph{Without refinement, \modelname{} stays ahead of the Transformer baselines.}
Prefilling without states and decoding with them keeps a relative gain of \(37\)--\(51\%\), at the prefill cost of a standard Transformer. In this mode \modelname{} is ahead of the token-matched Transformer at every scale, as \modelname{} w/o states already is, and it is ahead of the compute-matched Transformer at 350M and level with it at 1B (Figure~\ref{fig:prefill}). On arithmetic this mode keeps less of the gain, \(18\)--\(26\%\); it is still ahead of the compute-matched Transformer at every scale and of the token-matched one at 135M and 350M, and level with the latter at 1B.

\paragraph{Cost.}
Prefill without states has no sequential dependency along the prompt: it is one parallel pass. With \(R\) refinements, positions depend on each other only across consecutive passes, so the prompt is read in \(R+1\) parallel passes at \(R+1\) times the FLOPs of one. Sequential prefill costs the FLOPs of one pass, since the key-value cache avoids recomputation, but reads a prompt of \(n\) tokens in \(n\) dependent steps. A refined cache can thus be obtained either with one extra parallel pass, doubling the prefill FLOPs, or with \(n\) sequential steps at the FLOPs of one pass. Counting every forward pass, over the prompt or over one answer token, as one step, an answer of \(D\) tokens takes \(D+1\) steps without refinement and \(D+2\) with one: the extra pass adds \(1/(D+1)\) of the steps, a share that shrinks as the answer grows. At the mean answer lengths of our generative tasks, it adds \(17\%\) on TriviaQA (5 tokens), \(13\%\) on the generative QA suite (\(6.5\) tokens) and \(0.5\%\) on chain-of-thought BBH (209 tokens).

\begin{center}
\begin{minipage}{\linewidth}
\captionof{table}{Prefill modes of \modelname{} on arithmetic and the OLMES groups of Table~\ref{tab:lm}, with the token- and compute-matched Transformers of that table. Prefill cost: forward passes to read a prompt of \(n\) tokens, and FLOPs relative to one forward pass over the prompt. Full propagation: every answer token is fed the state predicted at the preceding position; --: a model without states. The prefill, decoding and cost columns describe how each model runs in deployment, where the answer is generated one token at a time. They are not the cost of scoring these benchmarks, where a given answer, as on MC and RC, is scored in one parallel pass by the models without states.}
\label{tab:prefill}
\small
\setlength{\tabcolsep}{4pt}
\resizebox{\linewidth}{!}{%
\begin{tabular}{@{}clllrrcccc@{}}
\toprule
 & & & & \multicolumn{2}{c}{Prefill cost} & & & & \\
\cmidrule(lr){5-6}
 & Model & Prefill & Decoding & passes & FLOPs & Arith.\ {\scriptsize BPB} $\downarrow$ & MC acc.\ \% & Gen.\ \% & RC {\scriptsize BPB} $\downarrow$ \\
\midrule
\multirow{8}{*}{\rotatebox[origin=c]{90}{135M}}
 & \modelname{} & w/o states & w/o states & 1 & 1\(\times\) & 2.42 & 38.7 & 12.0 & 1.105 \\
 & \modelname{} & w/o states & full propagation & 1 & 1\(\times\) & 2.40 & 38.9 & 13.9 & 1.093 \\
 & \modelname{} & 1 refinement & full propagation & 2 & 2\(\times\) & 2.30 & 39.3 & 15.3 & 1.079 \\
 & \modelname{} & 3 refinements & full propagation & 4 & 4\(\times\) & 2.29 & 39.4 & 15.5 & 1.078 \\
 & \modelname{} & 5 refinements & full propagation & 6 & 6\(\times\) & 2.29 & 39.4 & 15.5 & 1.078 \\
 & \modelname{} & sequential & full propagation & \(n\) & 1\(\times\) & 2.29 & 39.4 & 15.5 & 1.078 \\
 & Transformer, token-matched & -- & -- & 1 & 1\(\times\) & 2.46 & 38.1 & 13.6 & 1.110 \\
 & Transformer, compute-matched & -- & -- & 1 & 1\(\times\) & 2.50 & 39.5 & 14.1 & 1.080 \\
\midrule
\multirow{8}{*}{\rotatebox[origin=c]{90}{350M}}
 & \modelname{} & w/o states & w/o states & 1 & 1\(\times\) & 2.21 & 44.2 & 21.0 & 0.945 \\
 & \modelname{} & w/o states & full propagation & 1 & 1\(\times\) & 2.19 & 44.9 & 21.7 & 0.936 \\
 & \modelname{} & 1 refinement & full propagation & 2 & 2\(\times\) & 2.12 & 45.0 & 23.2 & 0.922 \\
 & \modelname{} & 3 refinements & full propagation & 4 & 4\(\times\) & 2.12 & 45.1 & 23.1 & 0.921 \\
 & \modelname{} & 5 refinements & full propagation & 6 & 6\(\times\) & 2.12 & 45.0 & 23.0 & 0.922 \\
 & \modelname{} & sequential & full propagation & \(n\) & 1\(\times\) & 2.12 & 45.0 & 23.2 & 0.921 \\
 & Transformer, token-matched & -- & -- & 1 & 1\(\times\) & 2.46 & 43.2 & 21.0 & 0.956 \\
 & Transformer, compute-matched & -- & -- & 1 & 1\(\times\) & 2.33 & 44.1 & 22.0 & 0.942 \\
\midrule
\multirow{8}{*}{\rotatebox[origin=c]{90}{1B}}
 & \modelname{} & w/o states & w/o states & 1 & 1\(\times\) & 2.05 & 51.2 & 32.9 & 0.791 \\
 & \modelname{} & w/o states & full propagation & 1 & 1\(\times\) & 2.03 & 51.6 & 33.9 & 0.784 \\
 & \modelname{} & 1 refinement & full propagation & 2 & 2\(\times\) & 2.00 & 52.5 & 35.2 & 0.772 \\
 & \modelname{} & 3 refinements & full propagation & 4 & 4\(\times\) & 1.99 & 52.5 & 35.1 & 0.772 \\
 & \modelname{} & 5 refinements & full propagation & 6 & 6\(\times\) & 1.99 & 52.5 & 35.2 & 0.772 \\
 & \modelname{} & sequential & full propagation & \(n\) & 1\(\times\) & 1.99 & 52.5 & 35.0 & 0.772 \\
 & Transformer, token-matched & -- & -- & 1 & 1\(\times\) & 2.03 & 50.5 & 32.9 & 0.808 \\
 & Transformer, compute-matched & -- & -- & 1 & 1\(\times\) & 2.07 & 51.6 & 34.8 & 0.792 \\
\bottomrule
\end{tabular}}
\end{minipage}
\end{center}

\FloatBarrier

\section{Cost Analysis}
\label{app:cost}

\subsection{Training and Inference FLOPs}
\label{app:overhead}

Table~\ref{tab:overhead} itemizes every addition \modelname{} makes to a vanilla Transformer's compute, excluding the teacher's forward pass, for the \(1\)B model (\(d=2048\), \(L=16\), \(N=1.07\)B non-embedding parameters, \(k=1024\), \(|\mathcal V|=100{,}352\)). A multiply-add counts as two FLOPs. Percentages are relative to the standard \(6N\) FLOPs per training token (\(2N\) forward, \(4N\) backward) over non-embedding parameters; counting the output head and the attention scores at \(4\)k context lowers every percentage by about a quarter. The scaling column uses \(N\approx16Ld^2\), the OLMo~2 block shape at \(1\)B (\(N\approx20Ld^2\) at \(32\)B, where the fusion block is \(0.85\%\)).

\begin{table}[ht]
\centering
\footnotesize
\setlength{\tabcolsep}{3pt}
\caption{Compute added per token by \modelname{} at \(1\)B, excluding the teacher forward. The trunk is the first \(90\%\) of training; the adaptation phase is the last \(10\%\) (\S\ref{sec:training}).}
\label{tab:overhead}
\begin{tabular}{@{}>{\raggedright\arraybackslash}p{4.0cm}lr>{\raggedleft\arraybackslash}p{1.75cm}>{\raggedright\arraybackslash}p{2.85cm}@{}}
\toprule
Component & FLOPs / token & \begin{tabular}[b]{@{}r@{}}\(1\)B\\(MFLOPs)\end{tabular} & \% of a \(6N\) step & Scaling \\
\midrule
Fusion block, Eq.~(\ref{eq:fusion}), forward + backward & \(66d^2\) & \(277\) & \(4.3\%\) & \(11d^2/N = 11/(16L)\), falls with depth \\
Soft-token sum \(E^{\top}\mathbf{s}_i\), forward + backward & \(4kd\) & \(8\) & \(0.13\%\) & \(k/(24Ld)\) \\
Forward KL on the top-\(k\) support + tail, forward + backward & \(\approx5|\mathcal V|+20k\) & \(0.5\) & \(<0.01\%\) & \(\approx5|\mathcal V|/(96Ld^2)\) \\
Prefix state dropout, learned bias & \(0\) & \(0\) & \(0\) & -- \\
\midrule
Trunk total, per training token & & \(286\) & \(4.4\%\) & \\
Adaptation phase: \(1.5\) extra no-gradient forwards on average & \(1.5\,(2N + 22d^2 + 2kd)\) & \(3{,}366\) & \(+52\%\) of those steps & fixed by the schedule \\
Whole run (\(0.9\times\) trunk \(+\ 0.1\times\) adaptation) & & & \(\approx9.6\%\) & \\
\midrule
Inference, per generated token & \(22d^2 + 2kd\) + top-\(k\) & \(97\) & \(4.5\%\) of a \(2N\) forward & \(11/(16L)\) \\
\bottomrule
\end{tabular}
\end{table}

Three remarks. First, the KL term needs the student's full-vocabulary log-partition at temperature \(\tau\), one extra pass over logits that the language-modeling loss already materializes; no \((B,T,|\mathcal V|)\) tensor is created beyond them. Second, the adaptation passes run without gradient and their fed-back distribution is detached, so they hold no activation memory; a step in that phase costs about \(1.5\times\) a trunk step. Third, over a whole run the additions total under \(10\%\) of the vanilla training FLOPs, and every term except the schedule-fixed adaptation shrinks with model size.

\paragraph{Compute-matched budgets.}
\label{app:compute-matched}
The compute-matched Transformers receive the training FLOPs of the whole \modelname{} run, including the teacher's forward pass, the fusion layer and the extra passes of the adaptation phase: 19.3B, 49.9B and 152.4B tokens at 135M, 350M and 1B, i.e., \(1.42\)--\(1.44\times\) \modelname{}'s budget. This accounting also charges a teacher forward pass during the adaptation phase, which our implementation mistakenly ran although its output is not used there (\S\ref{app:adaptation}); without it, the matched budgets would be 18.8B, 48.7B and 148.8B tokens. The compute-matched baselines thus train on about \(2.5\%\) more tokens than the method requires, which favors them.

\subsection{Memory Footprint}
\label{app:jacobi-footprint}

Table~\ref{tab:jacobi-footprint} reports the peak GPU memory of \modelname{}, T\textsuperscript{2}MLR and the Multi-pass Transformer in the setting of \S\ref{sec:jacobi}, measured with the same code, in the same container and on the same hardware. Setup: 2 NVIDIA B200 GPUs (183\,GB each) with data parallelism, PyTorch 2.9 (NVIDIA container 25.10), bf16, sequences of 2{,}048 tokens, a global batch of 256 sequences reached by gradient accumulation, and the production configuration of each method: T\textsuperscript{2}MLR(13,18) with 16 forward and 4 backward refinement passes, and \modelname{} in its trunk configuration, including the teacher forward pass and the top-\(k\) (\(k=1024\)) state construction, with the forward-KL term enabled. The latter makes this a conservative memory comparison for the KL-free model of \S\ref{sec:jacobi}. Peak memory is the maximum memory allocated by PyTorch on one GPU.

\begin{table}[ht]
\centering
\small
\setlength{\tabcolsep}{9pt}
\caption{Peak memory per GPU of \modelname{}, T\textsuperscript{2}MLR and the Multi-pass Transformer at 135M (SmolLM2 backbone, 2{,}048-token sequences, B200). Memory is the peak allocated by PyTorch. The Multi-pass Transformer's peak is measured on a three-pass batch, which determines the memory needed to run its full schedule.}
\label{tab:jacobi-footprint}
\begin{tabular}{@{}lcc@{}}
\toprule
Model & \begin{tabular}[c]{@{}c@{}}Micro-batch\\/ GPU\end{tabular} & Peak memory \\
\midrule
\modelname{} & 8 & \textbf{35.4 GB} \\
T\textsuperscript{2}MLR(13,18) & 8 & 44.7 GB \\
Multi-pass Transformer & 8 & 62.9 GB \\
\midrule
\modelname{} & 4 & \textbf{18.5 GB} \\
T\textsuperscript{2}MLR(13,18) & 4 & 23.0 GB \\
Multi-pass Transformer & 4 & 32.1 GB \\
\bottomrule
\end{tabular}
\end{table}

\modelname{}'s memory beyond a vanilla step includes the teacher's logits for the top-\(k\) selection and the dense soft-token matrix, both transient. T\textsuperscript{2}MLR stores the recurrent cache and the activations of the refinement passes it differentiates through. The Multi-pass Transformer has the largest peak memory because its three-pass batches retain activations from all three passes.

\FloatBarrier

\section{Extended Related Work}
\label{app:related}

\paragraph{Depth and sequential computation.}
Looped and recurrent-depth Transformers reapply a block of layers to increase the computation spent on each token~\citep{dehghani2019universal,saunshi2025reasoning,geiping2025scaling}; their inference cost grows with the number of iterations, and the depth available to a token is still bounded by the iterations run at that step. Chain of thought instead adds serial computation through generated tokens~\citep{wei2022chain}, which extends the expressivity of fixed-depth Transformers~\citep{feng2023towards,li2024serialcot,merrill2024cot}, and test-time methods scale it further through sampling, verification and search~\citep{wang2023selfconsistency,lightman2024lets,snell2025scaling}. Every such step, however, passes through the single decoded token. In \modelname{}, the computation's depth grows with the sequence at a fixed per-token cost (\S\ref{app:overhead}), and the approach is complementary to both.

\paragraph{Teachers as targets and as inputs.}
Pretrained teachers usually serve as targets: knowledge distillation matches their distributions~\citep{hinton2015distilling,kim2016sequence}, also in LM pretraining~\citep{gemmateam2024gemma2}, and richer objectives supervise future tokens or latents~\citep{gloeckle2024multitoken,zhang2026nitp}, leaving inference unchanged. An LM's distribution has also been fed to another model as an input, but one that remains at inference~\citep{sriram2018cold}. In \modelname{}, the teacher's distribution is a training input that the model learns to replace with its own, and our distillation baseline isolates the target role (\S\ref{sec:setting}). As with born-again and weak-to-strong students~\citep{furlanello2018born,burns2024weak}, the student can surpass its teacher (\S\ref{sec:synthetic}); our adaptation phase is scheduled sampling~\citep{bengio2015scheduled,mihaylova2019scheduled} on the state channel.

\paragraph{Comparison with the closest methods.}
Table~\ref{tab:related} lists, for the methods closest to \modelname{}, what is fed back, where it enters the model, whether the sampled token is kept, how the fed-back states are obtained during training, and at which stage the channel is learned. \modelname{} is the only method whose training-time states are supplied by an external pretrained LM, which is what allows a single parallel pass during pretraining.

\begin{table}[ht]
\centering
\scriptsize
\setlength{\tabcolsep}{2.5pt}
\renewcommand{\arraystretch}{1.15}
\caption{Methods that feed information back across generation steps. \emph{Training-time states}: how the fed-back states are obtained during training; \emph{own} means generated by the model being trained. $^\dagger$arXiv preprint.}
\label{tab:related}
\begin{tabular}{@{}>{\raggedright\arraybackslash}p{2.6cm}>{\raggedright\arraybackslash}p{2.75cm}>{\raggedright\arraybackslash}p{2.2cm}c>{\raggedright\arraybackslash}p{2.5cm}>{\raggedright\arraybackslash}p{1.45cm}@{}}
\toprule
Method & Fed-back signal & Enters at & Token & Training-time states & Stage \\
\midrule
Feedback Transformer~\citep{fan2021feedback} & All layers' states of past positions & Attention of every layer & kept & own, sequential & pretraining \\
RMT~\citep{bulatov2022recurrent} & Output memory tokens of the previous segment & Input of the next segment & kept & own, sequential over segments & pretraining \\
LCKV~\citep{wu2024lckv} & Top-layer keys and values & Attention of every layer & kept & own, iterative parallel passes & pretraining \\
Turbo Connection~\citep{tang2026turbo} & Higher-layer states of the previous token & Lower layers & kept & own, sequential in groups & fine-tuning \\
PonderLM~\citep{zeng2026ponderlm} & Top-$k$ soft token of its own prediction & Input of the same position & kept & own, extra passes per position & pretraining \\
PonderLM-2~\citep{zeng2026ponderlm2} & Last hidden state & An added input position & kept & own, Jacobi passes & pretraining \\
T\textsuperscript{2}MLR$^\dagger$~\citep{cai2026t2mlr} & Middle-layer hidden state & An earlier middle layer & kept & own, Jacobi passes & pretraining \\
Full-bandwidth Transformer$^\dagger$~\citep{wang2026fullbandwidth} & Top-layer hidden state & Input, gated with the token & kept & own, one to three parallel passes & pretraining \\
Latent Recurrent Transformer$^\dagger$~\citep{huang2026latent} & Hidden state of a source layer & Attention and residual stream & kept & own, interleaved parallel training & pretraining \\
WhiteMatter$^\dagger$~\citep{zhang2026whitematter} & Keys and values of all layers & Attention of every layer & kept & own, fixed-point passes & pretraining \\
\midrule
Coconut~\citep{hao2025coconut} & Last hidden state & Input & replaced & own, sequential & post-training \\
CODI~\citep{shen2025codi} & Projected last hidden state & Input & replaced & own, sequential & post-training \\
PCCoT~\citep{wu2025pccot} & Last hidden state & Input & replaced & own, Jacobi passes & post-training \\
Soft Thinking, MoI~\citep{zhang2025soft,zhuang2025mixture} & Soft token of its own distribution & Input & replaced, mixed & none (training-free) & inference only \\
HRPO~\citep{yue2025hybrid} & Soft token of its own distribution & Input, gated with the token & kept & own, RL rollouts & post-training \\
\midrule
CoT2~\citep{gozeten2026continuous} & Soft token & Input & replaced & teacher-forced, from an oracle & task training \\
CoLaR~\citep{tan2025colar} & Compressed reasoning-token embeddings & Input & replaced & teacher-forced, from gold traces & post-training \\
Thinking States$^\dagger$~\citep{amos2026thinkingstates} & Compressed reasoning of the previous chunk & A shallow layer & kept & teacher-forced, from gold annotations & post-training \\
CoCoMix~\citep{tack2026llm} & Its own predicted concept vector & Interleaved with hidden states & kept & own, same pass & pretraining \\
SMT$^\dagger$~\citep{kumar2026pretraining} & RNN memory & RNN state & -- & teacher-forced, from a jointly trained encoder & pretraining (RNN) \\
\midrule
\modelname{} (ours) & Top-$k$ next-token distribution & Input, fused with the token & kept & teacher-forced, from a pretrained LM & pretraining \\
\bottomrule
\end{tabular}
\end{table}

\end{document}